\documentclass[10pt,twocolumn,letterpaper]{article}

\usepackage{cvm}
\usepackage{times}
\usepackage{epsfig}

\usepackage{amsmath, amssymb}
\usepackage{graphicx}
\usepackage{bm}

\usepackage{cuted}
\usepackage{algorithmicx}

\usepackage{multirow}

\usepackage{booktabs}

\usepackage{algorithm}
\usepackage{algpseudocode}

\usepackage{xparse}

\usepackage{pgffor}
\usepackage{etoolbox}

\usepackage{float}

\usepackage{enumitem}

\usepackage{colortbl}
\usepackage{caption}
\usepackage{subcaption}
\usepackage{xspace}

\usepackage{arydshln}

\definecolor{my_red}{RGB}{243, 181, 179}
\definecolor{my_orange}{RGB}{248, 218, 182}
\definecolor{my_yellow}{RGB}{255, 254, 185}
\newcommand{\best}{\cellcolor{my_red}}
\newcommand{\sbest}{\cellcolor{my_orange}}
\newcommand{\tbest}{\cellcolor{my_yellow}}

\newcommand{\red}[1]{{\color{red}#1}}
\newcommand{\blue}[1]{{\color{blue}#1}}

\let\oldref=\ref
\renewcommand{\ref}[1]{\textcolor{green}{\oldref{#1}}}

\let\oldcite=\cite
\renewcommand{\cite}[1]{\textcolor{cyan}{\oldcite{#1}}}

\usepackage[pagebackref=true,breaklinks=true,letterpaper=true,colorlinks,bookmarks=false]{hyperref}

\newcommand{\keywords}[1]{{\bf \emph{Keywords: #1}}}

\cvmfinalcopy 

\def\cvmPaperID{225} 

\ifcvmfinal\fi
\begin{document}

\title{MIFO: Learning and Synthesizing Multi-Instance from One Image}

\author{Kailun Su\\
CSSE \& Math, ShenZhen University\\
{\small{kaslensu@gmail.com}}
\and
Ziqi He\\
CSSE, ShenZhen University\\
{\small{2022152008@email.szu.edu.cn}}
\and
Xi Wang\\
Independent Researcher\\
{\small hytidel333@gmail.com}
\and
Zhou Yang\thanks{Corresponding author} \\
CSSE, ShenZhen University\\
{\small{zhouyangvcc@szu.edu.cn}}
}
\maketitle

\begin{abstract}
This paper proposes a method for precise learning and synthesizing multi-instance semantics from a single image. The difficulty of this problem lies in the limited training data, and it becomes even more challenging when the instances to be learned have similar semantics or appearance.
To address this, we propose a penalty-based attention optimization to disentangle similar semantics during the learning stage. 
Then, in the synthesis, we introduce and optimize box control in attention layers to further mitigate semantic leakage while precisely controlling the output layout. 
Experimental results demonstrate that our method achieves disentangled and high-quality semantic learning and synthesis, strikingly balancing editability and instance consistency. 
Our method remains robust when dealing with semantically or visually similar instances or rare-seen objects. 
The code is publicly available at \url{https://github.com/Kareneveve/MIFO}

\end{abstract}

\section{Introduction}

\begin{figure}[t]
    \centering
    \includegraphics[width=1\linewidth]{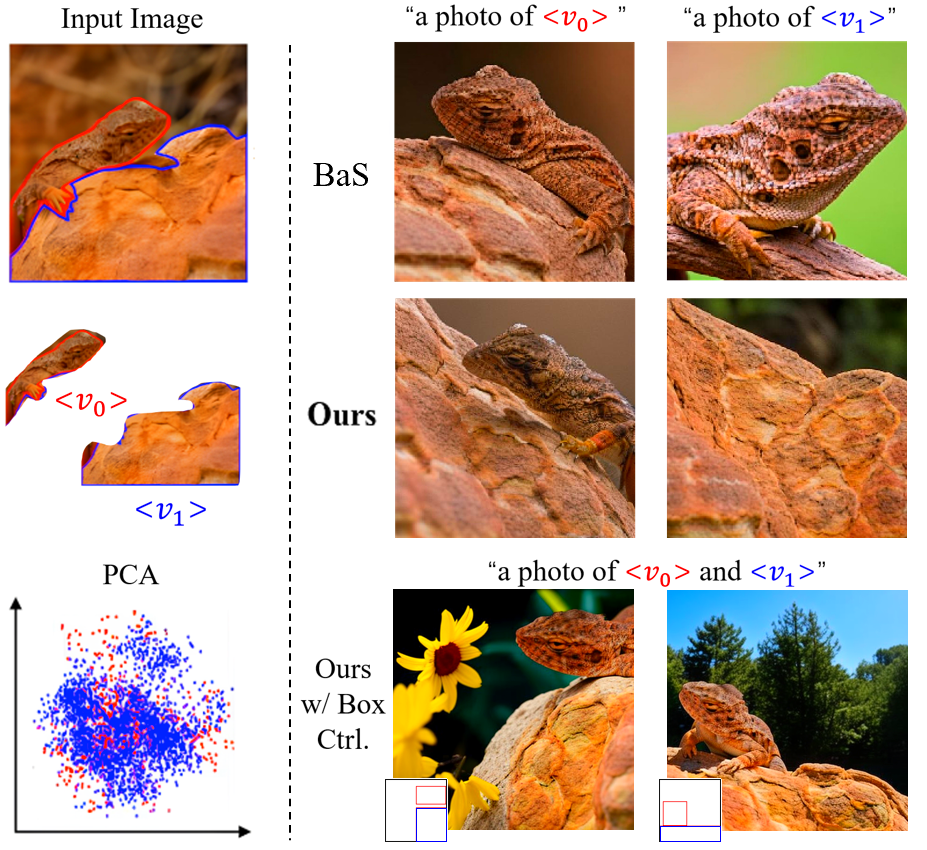}
    \caption{
        \textbf{Multi-object semantic learning for visually similar instances. } 
        Learning the semantics of multiple similar-looking instances from a single image is quite challenging, as their features are highly indistinguishable (left). Existing methods, such as Break-a-Scene (BaS)~\cite{avrahami2023break}, totally confuse the two objects (\red{$\langle  v_0 \rangle$} and \blue{$\langle v_1 \rangle$}) in synthesis (Row 1, right). Our method successfully disentangles these subtle features and produces correct synthesis that adheres to the prompts and additional box control (Rows 2 \& 3, right).
    }
    \label{fig-compare_BaS}
\end{figure}


Extracting semantic and appearance representations of objects from single real-world images plays a pivotal role in creative content generation and image editing~\cite{gal2022image,ruiz2023dreambooth,avrahami2023break,garibi2025tokenverse}. 
Recent success in diffusion models (DMs)~\cite{ho2020denoising,song2020denoising,ramesh2022hierarchical} 
brings breakthroughs into controllable image synthesis with instance consistency. 
For example, Textual Inversion (TI)~\cite{gal2022image} and DreamBooth (DB)~\cite{ruiz2023dreambooth} effectively learn the instance semantics from multiple reference images and reproduce the same instance in novel scenes. 
However, these approaches are limited to single-object learning from multiple samples, which is not always available in real-world applications, such as extracting and reconstructing multiple instances from a single image. 
This is essentially a more challenging problem: multi-instance semantic learning from a single example. 


A naive solution is to apply instance-level masks and learn each object separately. 
However, this often leads to degraded generation quality and very limited editability; see examples in our ablation study (Sec.~\ref{subsec-ablations}). 
To overcome this issue, Break-a-Scene (BaS)~\cite{avrahami2023break} proposes to learn the instances jointly, and meanwhile introduce a reward-based loss in cross-attention (CA) layers~\cite{vaswani2017attention} to encourage alignment between image semantics and text prompts. 
Though effective to some extent, BaS may fail when the instances share 
semantical or visual similarities, as shown in Fig.~\ref{fig-compare_BaS}. 
To understand this, we visualize the query features in the CA layers with principal components analysis (PCA). 
We can see that similar objects are significantly entangled in high-dimensional space, resulting in confusion and leakage in the synthesis. 



To effectively disentangle the semantics of different instances in a single image, 
We first identify that semantic leakage stems from the non-directional convergence states of the reward-based mechanism, where the optimization tends to converge to a mathematically optimal solution rather than the intended semantic target, as illustrated in Fig.~\ref{fig-reward_and_penalty_sample}.
To overcome this, we present a novel framework that consists of two stages: learning and synthesis. 
In the Disentangled Semantic Learning Stage, we incorporate reward-based attention control with penalty-based optimization to disentangle semantics in a coarse-to-fine manner. 
In the Precise Synthesis Stage, we introduce box control in both self-attention (SA) and CA layers to mitigate semantic fusion or leakage. 
Experiments demonstrate that, our method achieves disentangled and accurate multi-instance semantic learning and synthesis, yielding faithful and high-quality reconstruction and editing results. 
Our method shows excellent balance between editability and instance-consistency, and it remains robust when dealing with semantically or visually similar instances or rare-seen objects.

Our main contributions are summarized as follows: 


\begin{itemize}
    \item We reveal that the reward-based attention control used in prior works suffers from non-directional convergence, which fundamentally causes semantic leakage among visually or semantically similar instances.
    \item We propose a penalty-augmented attention optimization to complement the reward-based mechanism during semantic learning, enabling effective disentanglement of multi-instance semantics from a single image.
    \item In the Precise Synthesis stage, we enhance attention-layer box control with a hybrid in-box (reward-based) and out-of-box (penalty-based) formulation, which substantially mitigates semantic fusion and yields high-fidelity, instance-consistent compositions.
\end{itemize}


\section{Related Work}


\subsection{Semantic Learning}

Recently, many diffusion-based methods have been proposed to learn semantics from images. 
For example, Textual Inversion (TI)~\cite{gal2022image} optimized the text embedding to learn single-instance semantics from multiple samples for the first time. 
To further enhance the reconstruction fidelity, DreamBooth~\cite{ruiz2023dreambooth} and its variants~\cite{kumari2023multi,tewel2023key} fine-tuned the full diffusion U-Net, and even the text encoder. 
However, these methods require multiple reference images and are limited to single-instance learning. 
Another research avenue has turned to domain alignment. 
For example,~\cite{chen2023subject,wei2023elite} trained an image encoder to align text and image features in the latent space. 
Yet, it requires large-scale annotated text-image pairs and also fails to deal with rare-seen semantics due to limited data. 



Break-a-Scene (BaS)~\cite{avrahami2023break} explicitly learned multi-instance semantics from a single image by joint sampling (Appx.~\ref{appx_subsec-sampling_strategies}) with reward-based attention control. 
Though effective in most scenarios, the reward-based mechanism is inherently deficient when entanglement exists, \emph{i.e.}, it fails to distinguish semantics from different similar-looking instances due to semantic leakage (Sec.~\ref{semantic_learning}). 
To address this challenge, we introduce a {penalty-based} attention control to encourage semantic separation during the learning stage.


Apart from the above practice, recent research also turns to semantic learning in diffusion transformer~\cite{peebles2023scalable} architecture, \emph{e.g.}, TokenVerse~\cite{garibi2025tokenverse}. 
We omit further discussion here and instead focus on UNet-based approaches.


\begin{figure}[t]
    \centering
    \includegraphics[width=1\linewidth]{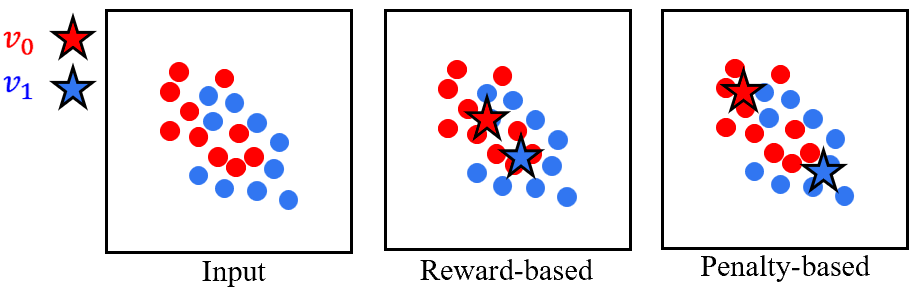}
    \caption{
        \textbf{Illustration of reward-/penalty-based attention control. } 
        Red and blue circles represent the query vectors in CA, while the stars denote the two tokens to optimize in semantic learning. As the features of \red{$\langle  v_0 \rangle$} and \blue{$\langle v_1 \rangle$}  are highly entangled, optimizing the tokens by considering only the positive samples (as reward-based approaches do) cannot distinguish the two objects. In contrast, our penalty-based solution aims to separate the tokens after semantic learning.
    }
    \label{fig-reward_and_penalty_sample}
\end{figure}

\begin{figure*}[t]  
    \centering
    \includegraphics[width=\textwidth]{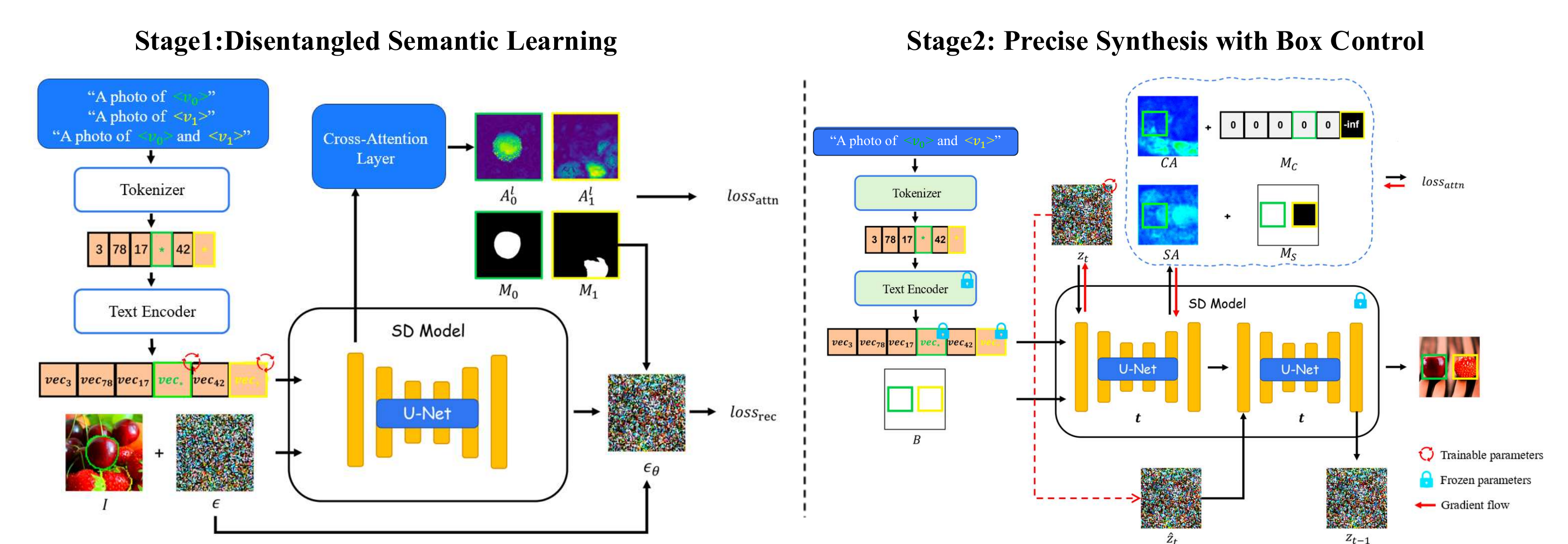}
    \caption{ \textbf{Framework of our method.} We divide the multi-instance semantic learning problem into two stages: Disentangled Semantic Learning for acquiring semantic and visual representations, and Precise Synthesis with Box Control for controlled reconstruction and synthesis. 
    Joint Sampling is employed in the semantic learning stage (see Appx.~\ref{appx_subsec-sampling_strategies}).}
    \label{fig-two_stage_pipeline}
\end{figure*}

\subsection{Precise Synthesis against Semantic Leakage}

Even if multiple similar semantics can be accurately separated and learned, applying them to instance-consistent image synthesis still faces challenges: the objects corresponding to semantic $\langle v_0 \rangle$ may also manifest visual features of $\langle v_1 \rangle$ (Fig.~\ref{fig-Sampling_Strategy_Ablation}), which means \textbf{semantic leakage}. 
Prior works~\cite{feng2022training,tunanyan2023multi,chefer2023attend} have attempted to strengthen semantic grounding but failed in multi-instance scenarios. 


\paragraph{Training-based Methods. }
Works like LayoutDiffusion~\cite{zheng2023layoutdiffusion}, Gligen~\cite{li2023gligen}, and Reco~\cite{yang2023reco} choose to train auxiliary conditional encoders to bind bounding boxes with text embeddings, thereby enhancing semantic grounding for pre-trained DMs. 
These methods rely on supervised training using image-box-label triplets from object detection datasets (\emph{e.g.}, COCO~\cite{lin2014microsoft}), resulting in limited generalization and reduced generation quality. 


\paragraph{Training-free Methods. } 
Some research turns to training-free methods~\cite{xie2023boxdiff,feng2022training,bar2023multidiffusion,dahary2024yourself,dahary2025decisive}.
For example, BoxDiff~\cite{xie2023boxdiff} and Chen \emph{et al.}~\cite{feng2022training} incorporate cross-attention mechanisms with box constraints for precise layout control and semantic grounding. 
However, inappropriate modification to attention weights may degrade generation quality. 
To overcome this issue, MultiDiffusion~\cite{bar2023multidiffusion} divides images into several regions and infers separately, followed by a region fusion mechanism. 
It successfully synthesizes high-quality content within each region, but suffers from unnatural blending across regions. 
Recently, Be Yourself~\cite{dahary2024yourself} proposed Bounded-Attention (BA) by performing attention-based clustering to refine the input regular boxes into irregular shapes, thereby enhancing generation quality. 

Nevertheless, for the above training-based or training-free methods, none of them have addressed the semantic leakage problem. 
In contrast, we incorporate box prompts for layout constraints and achieve precise and high-quality synthesis of multiple similar semantics via in-box and out-of-box control. 



\section{Method}

\subsection{Overview of Our Solution}
\label{subsec-overview_of_our_method}

To overcome the deficiency of reward-based 
attention 
control, we introduce a framework for learning and synthesizing multiple similar-looking instance semantics. 
As shown in Fig.~\ref{fig-two_stage_pipeline}, our solution contains two stages: 
i) Disentangled Semantic Learning Stage, where the input images and the instance-level masks user provided are fed into the DM for disentangled semantic learning, yielding text placeholders for target semantics; and ii) Precise Synthesis Stage, where users combine the learned embeddings with other prompting texts to achieve instance-consistent reconstruction and editing, while box control is introduced to alleviate semantic leakage or fusion. 


\begin{table*}[t]
    \centering
    \begin{tabular}{c | c | c}
        \hline
        \textbf{} & \textbf{In-box Control} & \textbf{Out-of-box Control} \\
        \hline
        \rule{0pt}{0.5ex} \\

        \parbox[t][-3.0cm][c]{0.1cm}{\rotatebox{90}{Self-Attention (SA)}} &
        \begin{subfigure}{0.4\textwidth}
            \centering
            \includegraphics[width=\linewidth]{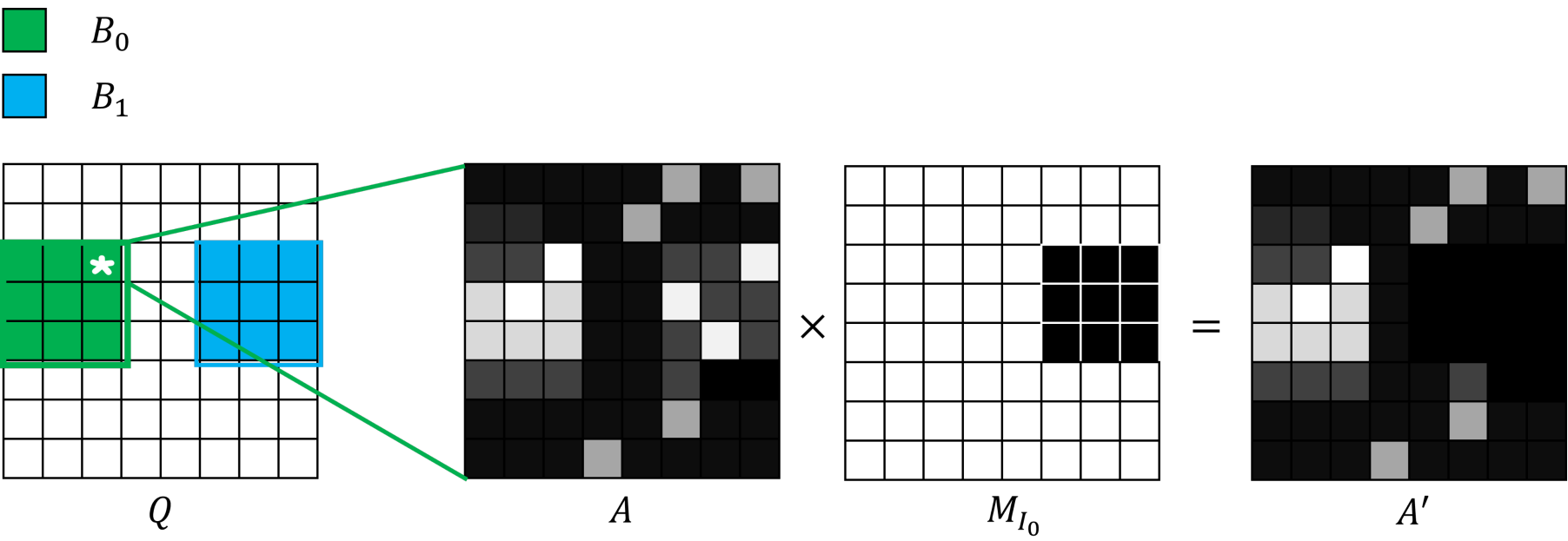}
            \subcaption*{(a) mitigate attention from leaking into other instances}
        \end{subfigure}
        &
        \begin{subfigure}{0.4\textwidth}
            \centering
            \includegraphics[width=\linewidth]{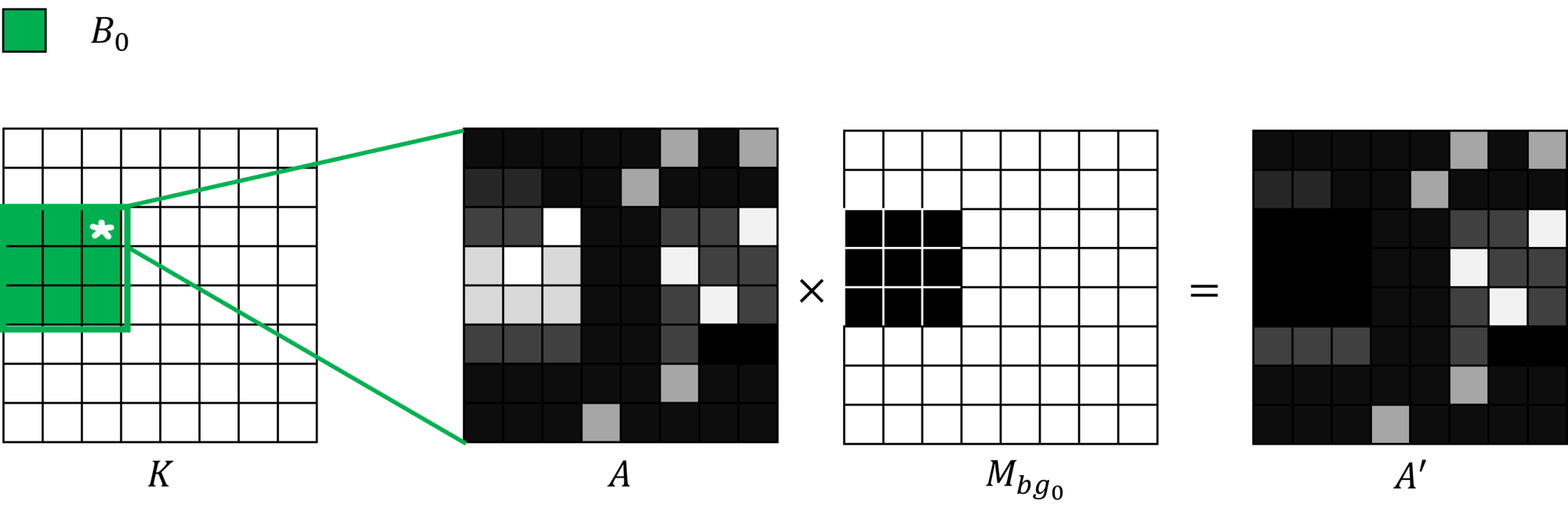}
            \subcaption*{(b) mitigate attention from leaking into the background}
        \end{subfigure} \\[0.5ex]  

        \parbox[t][-2.7cm][c]{0.1cm}{\rotatebox{90}{Cross-Attention (CA)}} &
        \begin{subfigure}{0.4\textwidth}
            \centering
            \includegraphics[width=\linewidth]{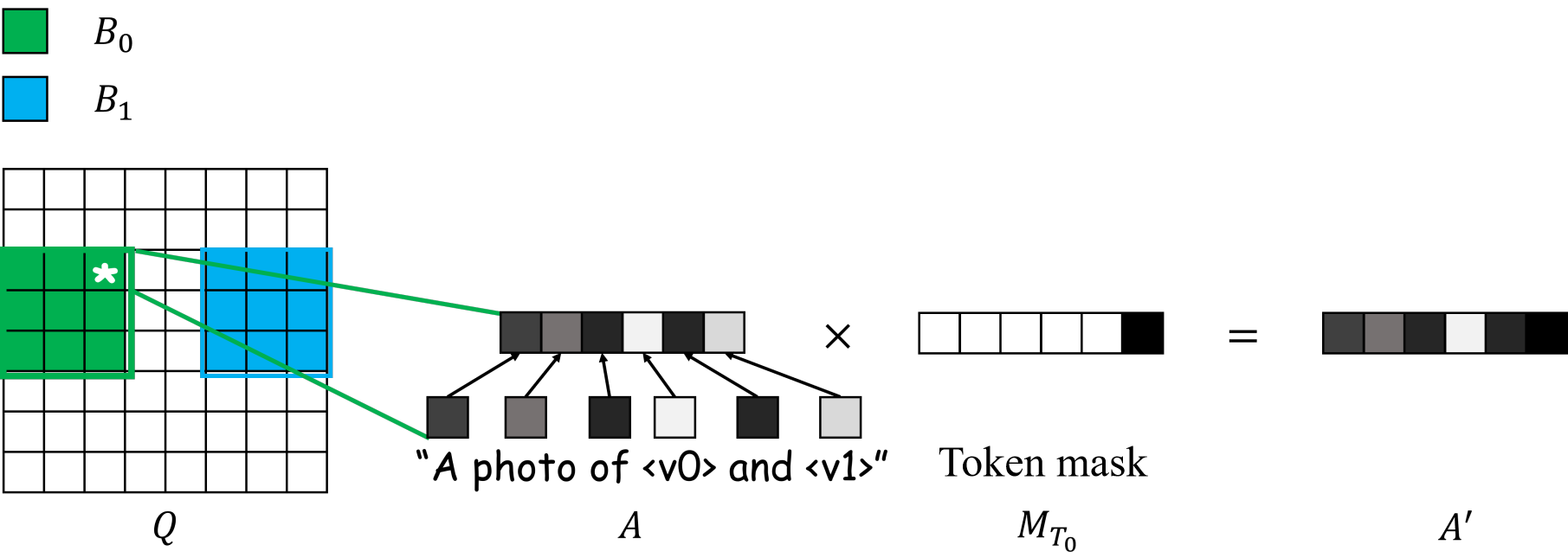}
            \subcaption*{(c) mitigate semantics from leaking into other instances}
        \end{subfigure}
        &
        \begin{subfigure}{0.4\textwidth}
            \centering
            \includegraphics[width=\linewidth]{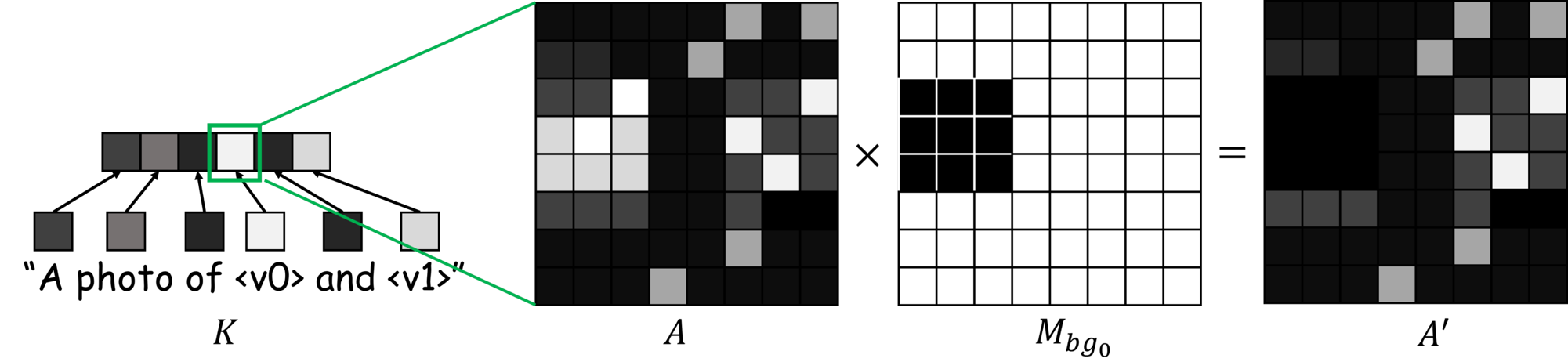}
            \subcaption*{(d) mitigate semantics from leaking into the background}
        \end{subfigure} \\

        \rule{0pt}{0.5ex} \\[-0.5ex]
        \hline
    \end{tabular}
    \captionof{figure}{
        \textbf{Illustration of in-/out-of-box control in Self-Attention (SA)/Cross-Attention (CA) layers. }
    }
    \label{fig-attn_control_grid_table}
\end{table*}

\subsection{Semantic Learning}
\label{semantic_learning}

Recent studies, such as reward-based attention control in BaS~\cite{avrahami2023break}, promote semantic disentanglement by strengthening the alignment between a specific text embedding and its corresponding image features (Fig.~\ref{fig-reward_and_penalty_sample}). This objective is achieved by minimizing  
\begin{equation}
    \mathcal{L}_{\text{CA}}^{\text{reward}} 
    = 
    \textstyle\sum\nolimits_{i = 0}^{N - 1} 
        \textstyle\sum\nolimits_{l \in \text{CA-Layers}} 
        \bigl\|
            \alpha M_i^l - A_i^l
        \bigr\|_2^2,
    \label{reward-base_loss_funtion}
\end{equation}
where $\alpha$ is a manually selected influence coefficient, $\text{CA-Layers}$ denotes the sampled cross-attention (CA) layers in the U-Net, and $M_i^l$ and $A_i^l$ represent the mask and the attention map associated with $\langle v_i\rangle$ in the $l$-th CA layer, respectively.  

Nevertheless, in high-dimensional latent spaces, the semantics of different instances often exhibit substantial entanglement (Fig.~\ref{fig-reward_and_penalty_sample}). During optimization, the reward-based loss predominantly emphasizes the semantics of the target instance while disregarding information from other co-located instances, thereby leading to semantic leakage. We point out that semantic leakage stems from the absence of discouraging misalignment, which results in non-directional convergence states (see Appx.~\ref{appx_sec-more_discussion_on_semantic_leakage} for more detailed analysis). 

To alleviate this issue and further disentangle semantically similar instances, we introduce a penalty-based attention loss applied to the CA layers. Unlike the reward-based loss, this penalty-based formulation suppresses correlations between a given text embedding and irrelevant image regions, while simultaneously promoting semantic separation across similar instances (see Fig.~\ref{fig-reward_and_penalty_sample}):  
\begin{equation}
    \mathcal{L}_{\text{CA}}^{\text{penalty}} 
    = 
    \textstyle\sum\nolimits_{i = 0}^{N - 1} 
        \textstyle\sum\nolimits_{l \in \text{CA-Layers}} 
        \bigl\|
            (\mathbf{1} - M_i^l) \odot A_i^l 
        \bigr\|_2^2,
    \label{penalty-base_loss_function}
\end{equation}
where $\mathbf{1}$ is a matrix of ones, $M_i^l$ is the mask of instance $i$ at layer $l$, and $A_i^l$ is the corresponding attention map.  

Since penalty-based methods are inherently sensitive to initialization (Sec.~\ref{subsec-ablations}), we adopt a coarse-to-fine optimization strategy to enhance training stability and performance. Specifically, we first employ the reward-based mechanism to guide text embeddings rapidly toward the region of entangled semantics. Subsequently, the penalty-based mechanism performs disentanglement and refinement, thereby enhancing the learning outcomes and effectively separating multiple instance semantics.

Formally, let $\Lambda = \{j_1, \cdots, j_k\}$ denote the index set of $k$ ($1 \leq k \leq N$) randomly selected target instances. The attention loss is then formulated as: 
\begin{small}
    \begin{equation}
        \mathcal{L}_{\text{attn}} =
        \begin{cases}
            \displaystyle
            \sum_{j \in \Lambda} \sum_{l \in \text{CA-Layers}} 
            \| \alpha M_{j}^l - A_{j}^l \|_2^2, 
            & e < e_{\text{coarse}}, \\[6pt]
            \displaystyle
            \sum_{j \in \Lambda} \sum_{l \in \text{CA-Layers}} 
            \| (\bm{1} - M_{j}^l) \odot A_{j}^l \|_2^2,
            & e \geq e_{\text{coarse}}.
        \end{cases}
    \end{equation}
\end{small} 
Here, $t$ is the timestep randomly sampled for the current iteration, $t_{\text{start}}$ is the starting timestep to incorporate the CA loss, $e$ denotes the current iteration index and $e_{\text{coarse}}$ is the iteration threshold required for coarse optimization.

Beyond the coarse-to-fine optimization scheme, we incorporate a Reconstruction Loss defined as  
\begin{equation}
\label{Eq-reconstruction_loss}
    \mathcal{L}_{\text{rec}} 
    = 
    \mathbb{E}_{\epsilon \sim \mathcal{N}(0, I)} 
        \Big[ 
            \| M_{\text{rec}} \odot \epsilon - M_{\text{rec}} \odot \epsilon_\phi(z_t, t, c_\theta) \|_2^2 
        \Big],
\end{equation}
where $\odot$ denotes the Hadamard product, and $M_{\text{rec}}$ corresponds to the union of masks for the selected instances (see Appx.~\ref{appx_sec-semantic_learning_details} for implementation details).  

The overall loss function in semantic learning is expressed as
\begin{equation}
    \mathcal{L} 
    = 
    \lambda_{\text{rec}} \mathcal{L}_{\text{rec}} 
    + 
    \lambda_{\text{attn}} \mathcal{L}_{\text{attn}}.
    \label{learning_stage_total_loss}
\end{equation}

Although optimizing text embeddings alone can partially achieve semantic alignment, the limited representational capacity of a single token often leads to suboptimal reconstruction fidelity. Following BaS~\cite{avrahami2023break}, we run DreamBooth~\cite{ruiz2023dreambooth} for a few hundred iterations after the semantic learning, which brings a more accurate appearance reconstruction. Specifically, we jointly fine-tune $\{\text{vec}_i\}$, the U-Net, and the text encoder. As shown in Fig.~\ref{fig-compare_BaS}, we can now synthesize new images of the learned instances using text prompts, resulting in high fidelity and synthesis quality. 


The pseudo-code of our disentangled semantic learning framework is summarized in Algorithm~\ref{alg-semantic_learning}.

\subsection{Synthesis Control}
\label{synthesis_control}

After acquiring the semantics and appearances of the selected instances, the next challenge is to achieve precise control over their synthesis.

Recently, BA~\cite{dahary2024yourself} introduced a reward-based In-Box control to enhance the localization of subject-specific semantics(Fig.~\ref{fig-attn_control_grid_table}). In particular, the loss encourages cross-attention maps to concentrate within the designated bounding boxes, thereby strengthening semantic-to-spatial alignment. When applied to attention layers, this reward further facilitates the separation of distinct subjects by constraining their interactions to their respective regions. Such a mechanism effectively mitigates semantic fusion and establishes clear subject-wise boundaries, laying the foundation for extending towards a complementary penalty-based strategy. The reward-based loss function is defined as follows:
\begin{equation}
\mathcal{L}_{\mathrm{attn}}^{\mathrm{fg},\ell}(i)
=
\sum_{j\in P_i}
\left\lVert
M_i \odot A^{\ell}_{i,j}
\right\rVert_2^2,
\end{equation}
\begin{equation}
\mathcal{L}_{\mathrm{attn}}^{\mathrm{bg},\ell}(i)
=
\sum_{j\in P_i}
\left\lVert
(\mathbf{1}-M_i)\odot A^{\ell}_{i,j}
\right\rVert_2^2,
\end{equation}
\begin{small}
\begin{equation}
\overline{\mathcal{L}}_{\mathrm{attn}}^{\mathrm{fg}}(i)
=\frac{1}{L}\sum_{\ell=1}^{L}\mathcal{L}_{\mathrm{attn}}^{\mathrm{fg},\ell}(i),
\ \ 
\overline{\mathcal{L}}_{\mathrm{attn}}^{\mathrm{bg}}(i)
=\frac{1}{L}\sum_{\ell=1}^{L}\mathcal{L}_{\mathrm{attn}}^{\mathrm{bg},\ell}(i),
\end{equation}
\end{small}
\begin{equation}
L_i^{\mathrm{reward}}
=
\left(1 \;-\;
\frac{\displaystyle \overline{\mathcal{L}}_{\mathrm{attn}}^{\mathrm{fg}}(i)}
{\displaystyle \overline{\mathcal{L}}_{\mathrm{attn}}^{\mathrm{fg}}(i)
+\overline{\mathcal{L}}_{\mathrm{attn}}^{\mathrm{bg}}(i)}\right)^2
\end{equation}

However, in the context of multi-instance semantic learning from a single image, relying solely on in-box control induces semantic leakage into the background, thereby hindering accurate instance reconstruction. A straightforward solution, as adopted in Be Decisive~\cite{dahary2025decisive}, is to treat the complement of the bounding boxes as an additional \((k+1)\)-th region for synthesis. For greater flexibility and methodological consistency, we instead extend BA~\cite{dahary2024yourself} by introducing a penalty-based out-of-box control term(Fig~\ref{fig-attn_control_grid_table}), which explicitly discourages semantic leakage into the background:
\begin{equation}
L_i^{\mathrm{penalty}}=\log\!\bigl(1+\overline{\mathcal{L}}_{\mathrm{attn}}^{\mathrm{bg}}(i)\bigr)
\end{equation}
Nonetheless, excessive penalization of semantic leakage may impair the coherence between instances and their surrounding background. To alleviate this issue, we adopt a dynamic decay strategy that progressively reduces the penalty weight. Specifically, the schedule first linearly decreases the weight to an intermediate value, followed by cosine annealing towards a minimal value. This two-stage design prevents over-penalization in the early phase while ensuring stable convergence and improved integration between instances and background:
\begin{equation}
\resizebox{0.95\columnwidth}{!}{$
\alpha(t) =
\begin{cases}
\alpha_{\max} + \dfrac{t-1}{S_1 - 1}\left(\alpha_{\min} - \alpha_{\max}\right),
& 1 \leq t \leq S_1, \\[10pt]
\alpha_{\text{final}} +
\dfrac{1 + \cos\!\left(\dfrac{\pi (t - S_1)}{N - S_1}\right)}{2}
\left(\alpha_{\min} - \alpha_{\text{final}}\right),
& S_1 < t \leq N
\end{cases}
$}
\label{eq-alpha_decay}
\end{equation}

In summary, we apply box control to both self-attention and cross-attention layers, combining reward and penalty terms with a time-dependent weight:
\begin{equation}
    \mathcal{L}_{\text{attn}}^{\text{SA}} = L_{i, \text{SA}}^{\mathrm{reward}} + \alpha(t)L_{i, \text{SA}}^{\mathrm{penalty}},
\end{equation}
\begin{equation}
    \mathcal{L}_{\text{attn}}^{\text{CA}} = L_{i, \text{CA}}^{\mathrm{reward}} + \alpha(t)L_{i, \text{CA}}^{\mathrm{penalty}}
\end{equation}
The final attention loss is thus defined as:
\begin{equation}
    \mathcal{L}_{\text{attn}} 
    = 
    \lambda_{\text{attn}} ^{\text{SA}} \mathcal{L}^{\text{SA}}_{\text{attn}} 
    + 
    \lambda_{\text{attn}}^{\text{CA}} \mathcal{L}^{\text{CA}}_{\text{attn}},
    \label{eq-combined_attn_loss}
\end{equation}
and the latent $z_t$ is optimized before each denoising step as:
\begin{equation}
    z_t^{\mathrm{opt}}=z_t-\beta\nabla_{z_t}\sum_i\mathcal{L}_i^2,
\end{equation}
thereby confining multi-instance semantics within their designated spatial and textual scopes.

After the aforementioned preliminary fusion optimization in the latent space, BA~\cite{dahary2024yourself} further points out that coarse masking in the later stage may degrade image quality and introduce boundary artifacts. To address this, we follow BA~\cite{dahary2024yourself} and replace each bounding box in the later stage with a fine-grained segmentation mask obtained by clustering self-attention maps.

The algorithm is shown in Algorithm~\ref{alg-semantic_synthesis}.

\section{Experiments}
\label{sec-experiments}

\subsection{Experimental Settings}

Experiments are conducted on 30 images of size 512x512, each containing at least two instances, with each instance occupying at least 15\% of the full image. 
Particularly, 15 images contain instances with high semantical or visual similarity, while others include semantically independent ones. 
We adopt the pre-trained Stable Diffusion V2.1~\cite{rombach2022high} as our base model. Text prompts are in the form of "\texttt{a photo of ...}" for both learning and synthesis. 

For the 1200 iterations of the disentangled semantic learning stage, the first 800 steps optimize only the text embeddings for semantic learning, while the remaining 400 steps jointly fine-tune the U-Net and the text encoder to learn the appearance (\emph{i.e.}, DreamBooth). 
The training process begins with 200 reward-based iterations, followed by 600 penalty-based steps. 
The entire learning stage costs $\sim$15 min per image. 


We employ 50-step DDIM~\cite{song2020denoising} sampling during the synthesis stage, where the first 15 steps apply both in-box and out-of-box control for latent optimization, while the remaining 35 steps retain only in-box constraints for attention refinement. 
Each image is generated in $\sim$8 min. 

\begin{figure}[t]
    \centering
    \includegraphics[width=1\linewidth]{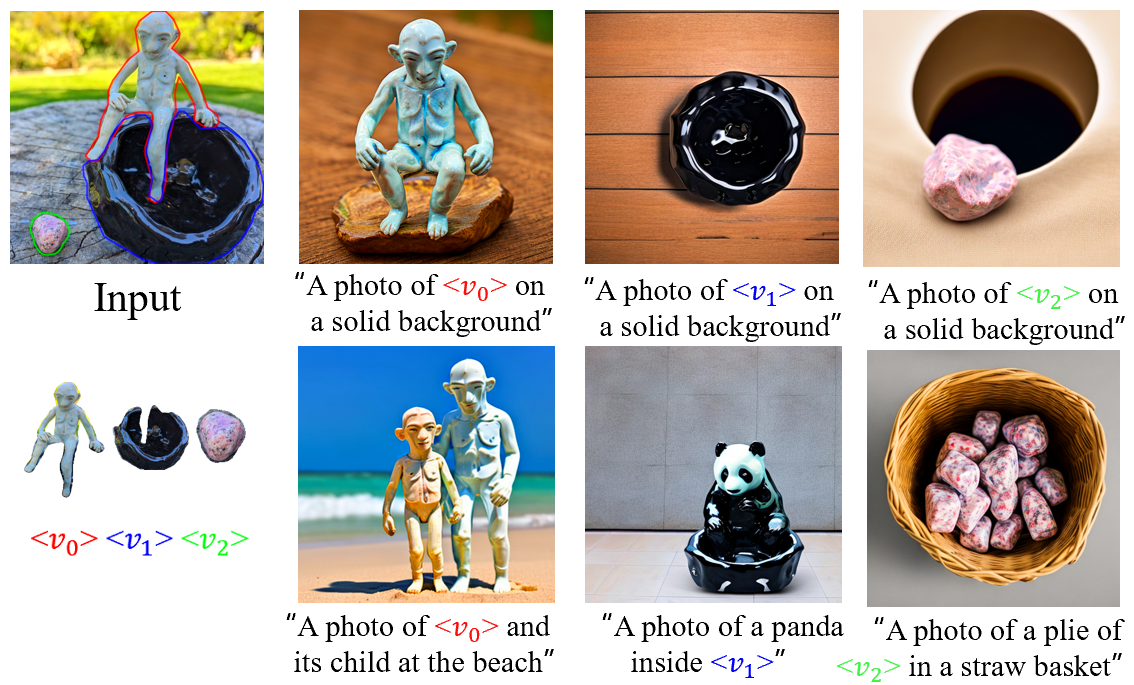}
    \caption{
        \textbf{Results of semantic learning and precise synthesis with rare-seen objects. }
    }
    \label{fig-mult_obj_learning_result_rare}
\end{figure}

Fig.~\ref{fig-mult_obj_learning_result_rare} presents a challenging types of multi-instance semantic learning problem, since rare-seen objects are difficult to describe with text. 
Results show that our method learns the semantics separately, and achieves precise synthesis and editing. 

Refer to Appx.~\ref{appx_sec-experimentall_details} for more details about experimental settings and evaluation.

\subsection{Multi-instance Semantic Learning}
\label{subsec-multi_instance_semantic_learning}

\paragraph{Baselines.} 
We benchmark our method against Textual Inversion (TI)~\cite{gal2022image}, DreamBooth (DB)~\cite{ruiz2023dreambooth}, and Break-a-Scene (BaS)~\cite{avrahami2023break}. 
Since TI and DB require multiple images for single-instance semantic learning and do not support segmentation input, we employ the joint sampling (Appx.~\ref{appx_subsec-sampling_strategies}) to augment the single reference image to a set of images, which is subsequently used as inputs. 
We name these two variants with masks as TI-m and DB-m, respectively. 
Besides, we adopt the public implementation and weights of BaS. 


\begin{figure}[t]
    \centering
    \includegraphics[width=1\linewidth]{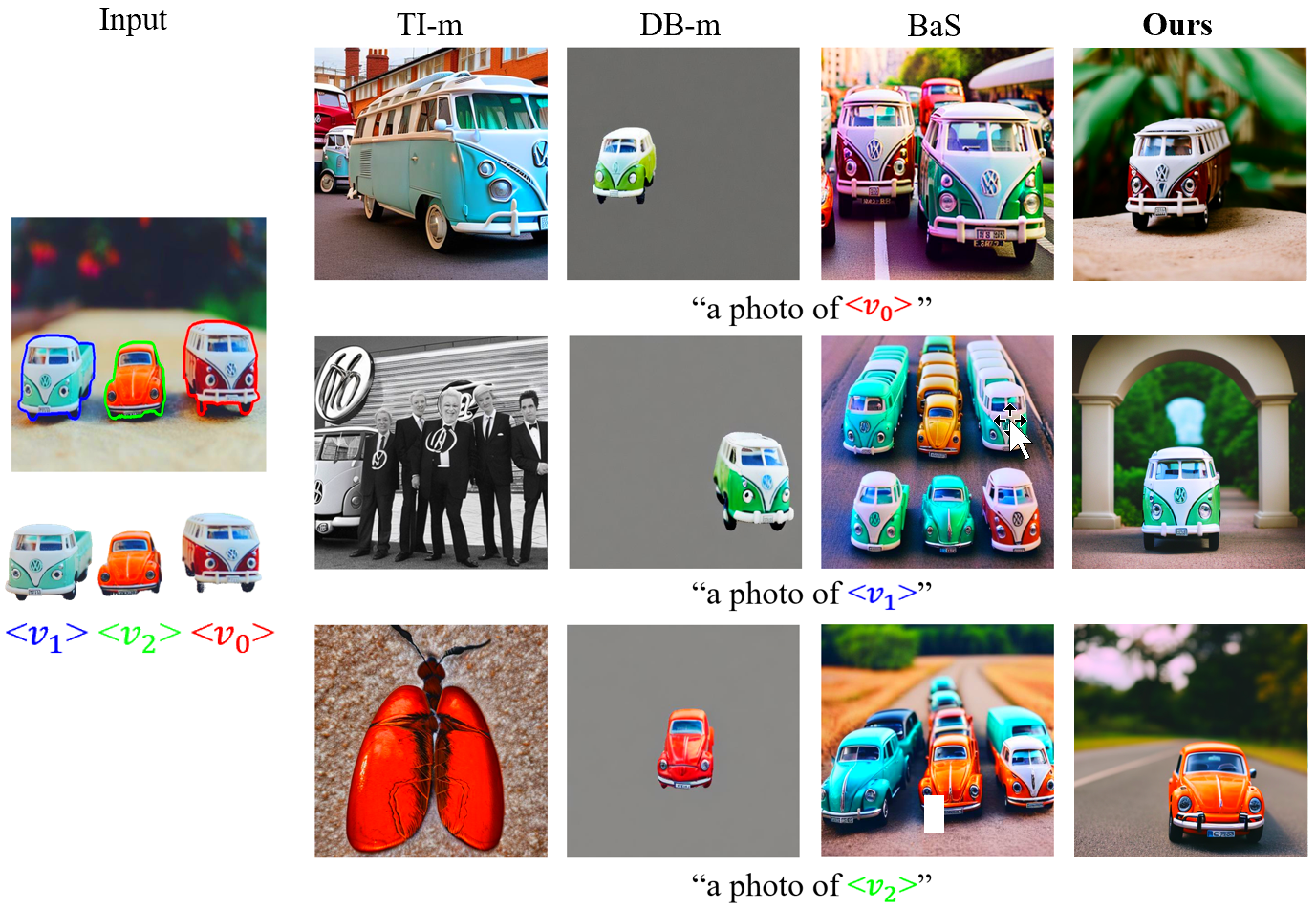}
    \caption{
        \textbf{Qualitative comparison of semantic learning. } 
    }
    \label{fig-Semantic_Learning_Comparison}
\end{figure}

\paragraph{Qualitative Comparison.} 
The results shown in Fig.~\ref{fig-Semantic_Learning_Comparison} demonstrate that TI-m fails to learn multiple instances, DM-m significantly overfits, and BaS struggles with semantic leakage. More qualitative results, such as images with semantic or visual similarities are shown in Fig.~\ref{fig-mult_obj_learning_result_semantic} and Fig.~\ref{fig-mult_obj_learning_result_visual} in Appx.~\ref{appx_subsec-more_qualitative_results}, respectively. 
In contrast, our method achieves accurate multi-instance learning and reasonable synthesis. 
Refer to Fig.~\ref{fig-more_qualitative_comparison_semantic_learning} in Appx.~\ref{appx_subsec-more_qualitative_results} for more qualitative comparison. 



\begin{figure}[t]
    \centering
    \includegraphics[width=1\linewidth]{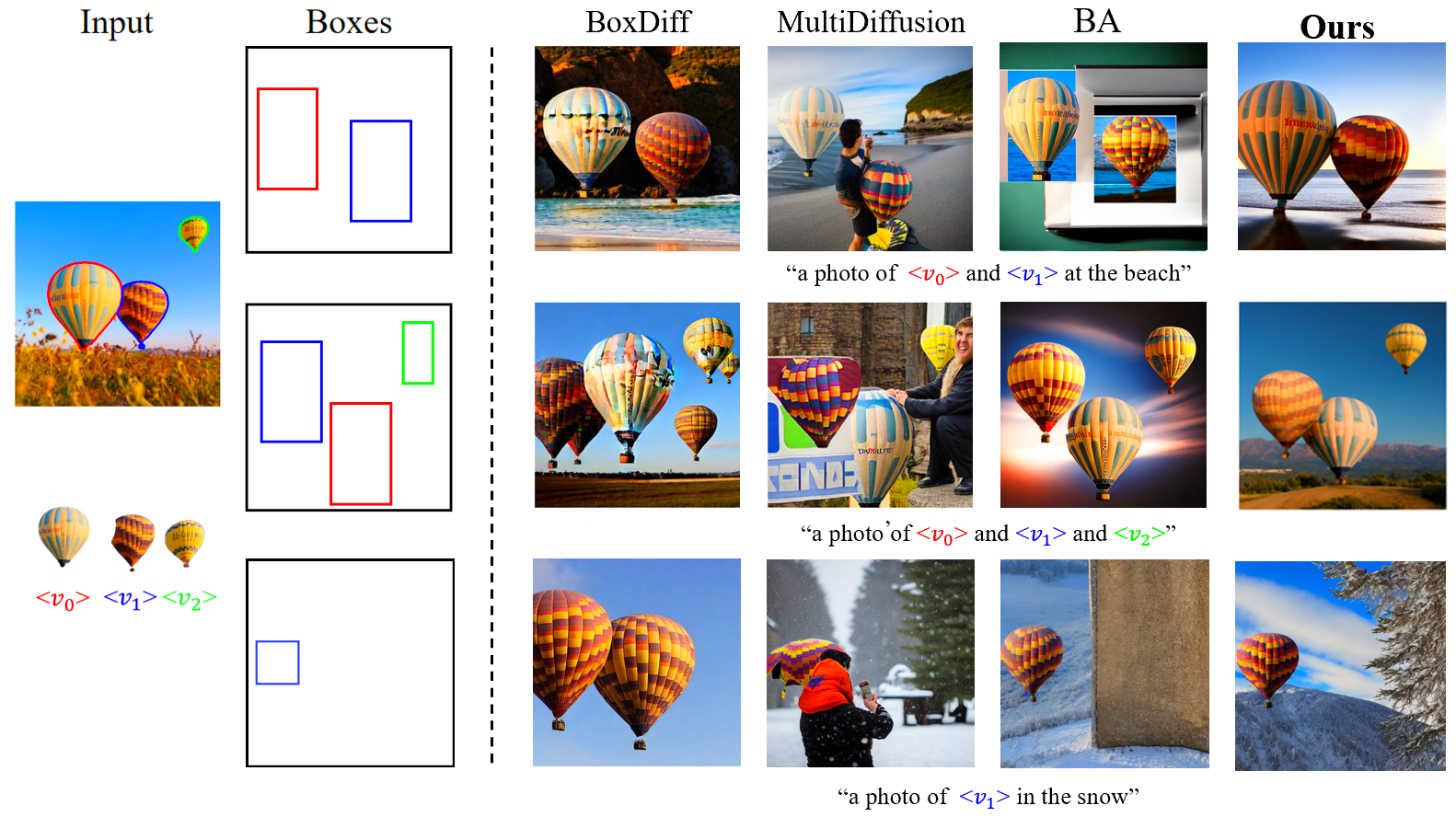}
    \caption{
        \textbf{Qualitative comparison of precise synthesis.} 
    }
    \label{fig-mult_obj_Semantic_Synthesis_Comparison}
\end{figure}

\begin{table}[t]
\centering
\caption{\textbf{Quantitative comparison on semantic metrics with composite score.}}
\label{tab-semantic_learning}
\small
\resizebox{0.9\columnwidth}{!}{%
\setlength{\tabcolsep}{4pt}
\begin{tabular}{lccc}
    \toprule
    \textbf{Method} &
    \textbf{SIM-D} $\uparrow$ &
    \textbf{NSIM-D} $\downarrow$ &
    \textbf{SIM-D $\times$ NSIM-D} $\uparrow$ \\
    \midrule
    TI-m         & 0.7525 & \best{0.7094} & 0.5338 \\
    DB-m         & \tbest{0.7685} & \sbest{0.7220} & \tbest{0.5548} \\
    BaS          & \sbest{0.7781} & \tbest{0.7253} & \sbest{0.5643} \\
    \textbf{Ours} & \best{0.7918} & 0.7258 & \best{0.5746} \\
    \bottomrule
\end{tabular}%
}
\end{table}

\begin{table}[t]
\centering
\caption{\textbf{Quantitative comparison} with transformed metrics and composite score. 
The last column shows the user preference ratio of the baselines (the left number) compared to our method (the right number) in user study (Appx.~\ref{appx-user_study}).}
\label{tab-semantic_synthesis}
\small
\resizebox{0.98\columnwidth}{!}{%
\setlength{\tabcolsep}{3pt}
\begin{tabular}{ccccccc}
    \toprule
    \textbf{Method} & 
    \textbf{SIM-C} $\uparrow$ & 
    \textbf{SIM-D} $\uparrow$ & 
    \textbf{NSIM-D} $\downarrow$ & 
    \textbf{CS} $\uparrow$ & 
    \textbf{HPS v2} $\uparrow$ &
    \textbf{User Preference}  \\
    \midrule
    BD      & \tbest{0.608} & 0.762 & \best{0.721} & 0.129 & \sbest{25.312} & 38.7\% vs. \textbf{61.3\%} \\
    MD      & \best{0.628} & \tbest{0.778} & 0.730 & \sbest{0.132} & \best{26.339} & 17.5\% vs. \textbf{82.5\%}\\
    BA      & 0.606 & \sbest{0.790} & \tbest{0.727} & \tbest{0.131} & 24.943 & 19.2\% vs. \textbf{80.8\%}\\
    \textbf{Ours} & \sbest{0.609} & \best{0.792} & \sbest{0.726} & \best{0.132} & \tbest{25.094} & /\ \\
    \bottomrule
\end{tabular}%
}
\end{table}

\paragraph{Quantitative Comparison.}
The comparison in Tab.~\ref{tab-semantic_learning} demonstrates that:
Our method achieves the highest level of instance consistency (SIM-D) while maintaining an excellent balance between editability and consistency, as reflected by the composite metric \(\mathrm{SIM\text{-}D}\times\mathrm{NSIM\text{-}D}\). In contrast, TI-m and DB-m adopt mask-based separate sampling but fail to effectively disentangle semantics across different instances, often producing results with blended visual features (see Sec.~\ref{subsec-ablations}). Although BaS demonstrates comparable performance on NSIM-D, our approach consistently outperforms it on SIM-D, which serves as the primary indicator for multi-instance semantic learning.

\subsection{Multi-instance Precise Synthesis}

\paragraph{Baselines.} 
We benchmark our method against BoxDiff (BD)~\cite{xie2023boxdiff}, MultiDiffusion (MD)~\cite{bar2023multidiffusion} and Bounded-Attention (BA)~\cite{dahary2024yourself}. 
Both of them support box control without re-training. 
We omit comparisons with GLIGEN~\cite{li2023gligen}, Attention-Refocusing~\cite{phung2024grounded} and ReCo~\cite{yang2023reco}, as their performance is inferior to that of BA~\cite{dahary2024yourself}.

\paragraph{Qualitative Comparison.}

As shown in Fig.~\ref{fig-mult_obj_Semantic_Synthesis_Comparison}, BD struggles with semantic leakage and artifacts, while MD alleviates leakage at the cost of reduced spatial coherence, and BA exhibits limited capability in background generation.
In contrast, our method yields composition with cleaner semantics and reasonable spatial coherence. 
Refer to Fig.~\ref{fig-more_qualitative_comparison_semantic_synthesis} in Appx.~\ref{appx_subsec-more_qualitative_results} for more comparisons. 





\paragraph{Quantitative Comparison.} 
The comparison in Tab.~\ref{tab-semantic_synthesis} demonstrates that:  
Both BA and our method adjust the attention weights of background pixels in attention layers to optimize the latent representation; however, this modification may inadvertently affect the semantic perception of other tokens and pixels, leading to weaker performance compared to alternative methods on HPS v2. Despite this, our method achieves the highest SIM-D score, underscoring its superior ability to preserve consistency between the synthesized result and the original instance. Although MD and BD obtain higher SIM-C scores, reflecting the limitations of our method in handling long prompts (see Appx.~\ref{Limitations}), our approach nonetheless demonstrates an excellent balance between appearance consistency, text consistency, and mitigation of semantic leakage between instances, as captured by the composite metric (CS, defined in Appx.~\ref{appx_subsec-evaluation_metrics}). Moreover, user preference studies further validate the effectiveness of our approach, showing that it outperforms all baselines with an overall winning rate above 60\% and surpasses 80\% against BA, thereby indicating a clear human preference for our results.

\subsection{Ablations}
\label{subsec-ablations}

\begin{figure}[t]
    \centering
    \includegraphics[width=1\linewidth]{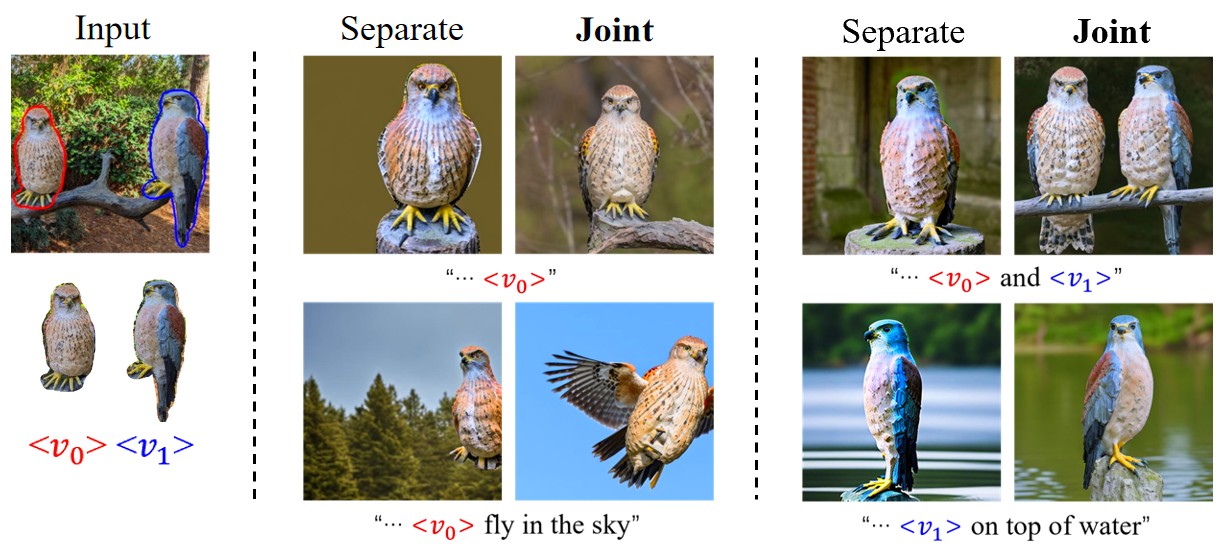}
    \caption{
        \textbf{Ablations on sampling strategies.}
        Separate Sampling isolates each instance for individual semantic learning, and Joint Sampling randomly combines multiple instance masks to enhance data diversity (see Appx.~\ref{appx_subsec-sampling_strategies}).
    }
    \label{fig-Sampling_Strategy_Ablation}
\end{figure}

\paragraph{Sampling Strategy.} We first compare separate and joint sampling in multi-instance semantic learning and synthesis. 
Fig.~\ref{fig-Sampling_Strategy_Ablation} presents the reconstruction (row 1) and editing (row 2) results of these strategies. 

For reconstruction, separate sampling in single-instance scenarios struggles with maintaining foreground-background coherence, thereby reducing reconstruction quality. In dual-instance settings, it further fails to distinguish semantics between the two instances, often producing results with blended visual features. By contrast, joint sampling demonstrates a stronger capacity to differentiate semantics between instances, yet it remains prone to semantic leakage during reconstruction. This limitation can be alleviated through the incorporation of box control, as discussed in Sec.~\ref{synthesis_control} of the main paper.

For editing, separate sampling yields results that misalign with text prompts, while joint sampling produces accurate and high-quality coherence with excellent text prompt alignment. 



In summary, joint sampling preserves semantic consistency and reduces artifacts, which is adopted in this study unless otherwise specified. 


\begin{figure}[t]
    \centering
    \includegraphics[width=1\linewidth]{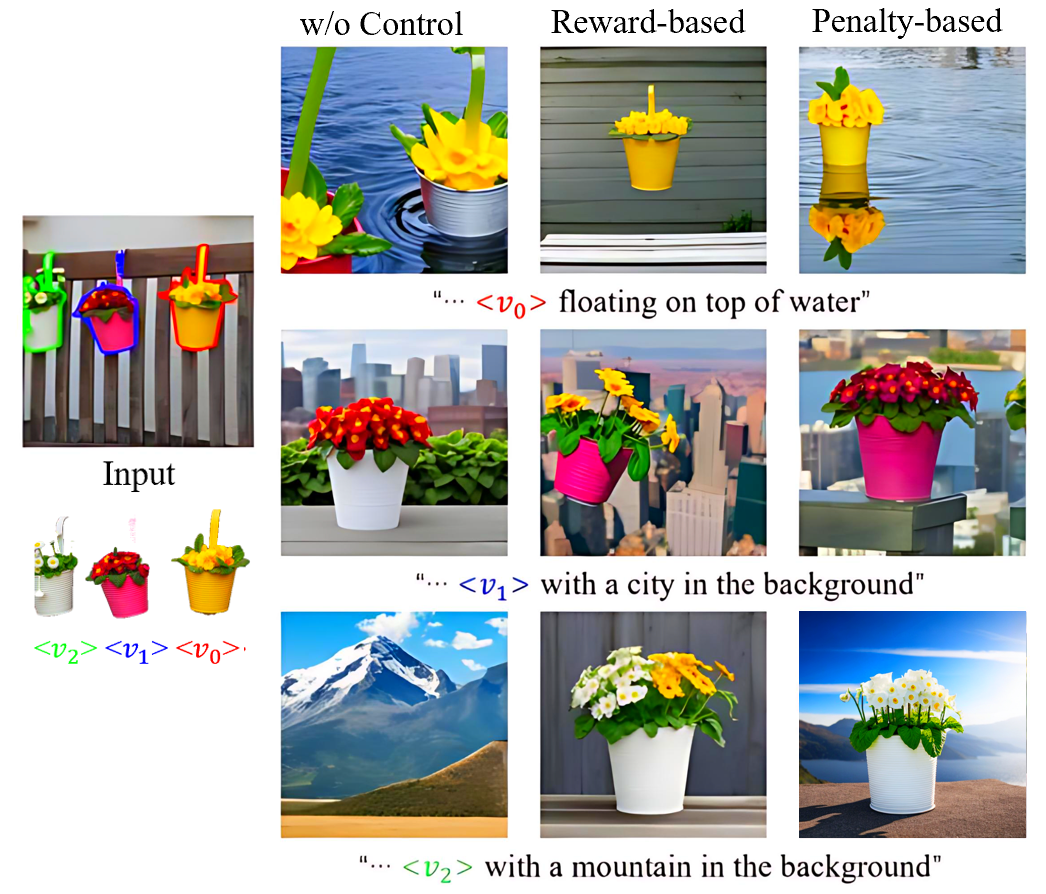}
    \caption{
        \textbf{Comparison between different control strategies in semantic learning. }
    }
    \label{fig-reward_and_penlty_result}
\end{figure}

\paragraph{Reward-/Penalty-based Semantic Learning.} 
We conduct ablations on different control strategies, and the results in Fig.~\ref{fig-reward_and_penlty_result} reveal distinct characteristics of each approach. The uncontrolled method fails to guide the text embeddings toward the corresponding semantics in multi-instance scenarios, resulting in blended visual features, as shown in rows 1 and 2, and semantic omission, as in row 3. The reward-based method achieves alignment between text embeddings and semantics, but it struggles to disentangle semantic correlations, which manifests in entangled visual features in rows 2 and 3. In contrast, our proposed penalty-based method enables accurate semantic disentanglement while preserving high editability.


\begin{figure}[t]
    \centering
    \includegraphics[width=1\linewidth]{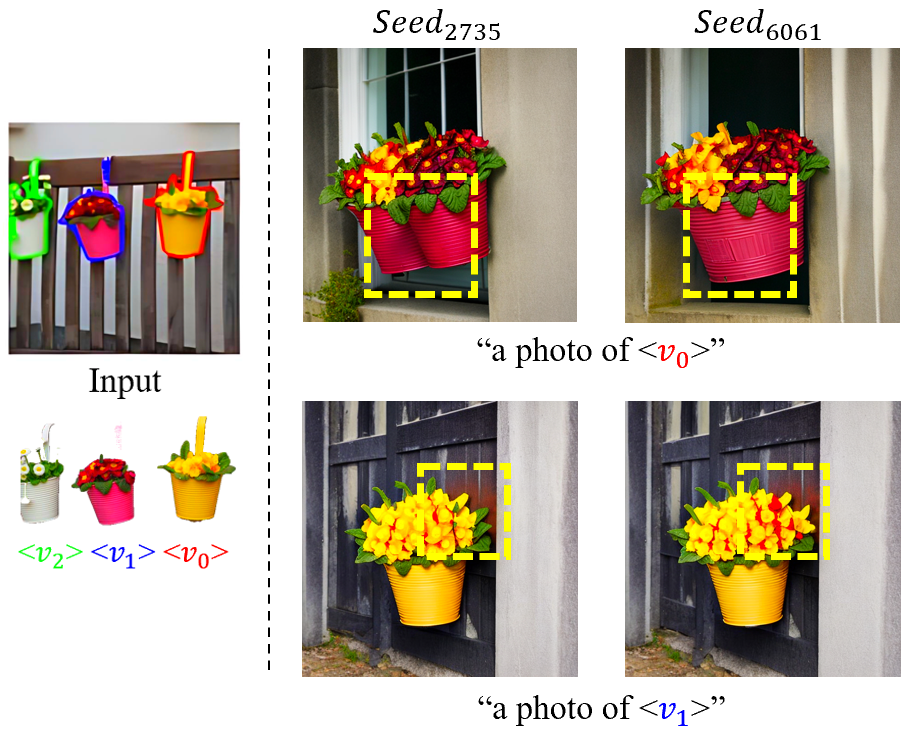}
    \caption{
    \textbf{Comparison between different initializations with the penalty-only method.}
    The red flowerpot (Row 1) is split into two instances under seed 2735, while seed 6061 (Row 2) amplifies semantic leakage from the red to the yellow flowers.
    }
    \label{fig-penalty_only_with_diff_seed}
\end{figure}

\paragraph{Only Penalty-based Semantic Learning.}
As noted in Sec.~\ref{synthesis_control}, penalty-based control alone is sensitive to initialization, with different seeds yielding inconsistent semantic learning and reconstruction, as shown in Fig.~\ref{fig-penalty_only_with_diff_seed}. This highlights the necessity of our two-stage optimization strategy.

\begin{figure}[htbp]
    \centering
    \includegraphics[width=1\linewidth]{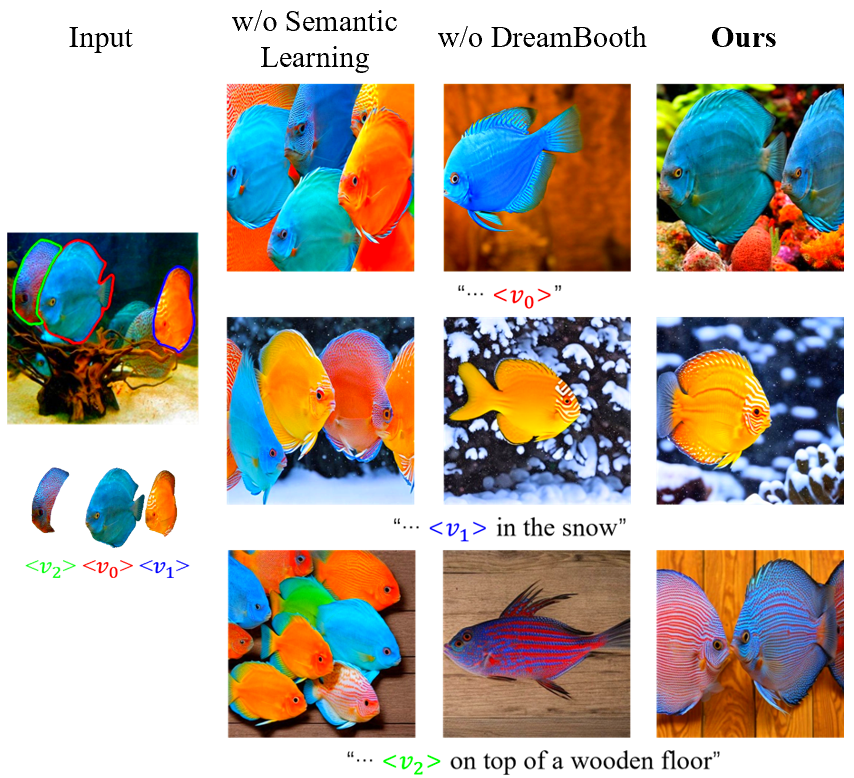}
    \caption{
        \textbf{Ablations on the steps in the disentangled semantic learning stage. }
    }
    \label{fig-Training_Phase_Ablation}
\end{figure}
\paragraph{Two-step Disentangled Semantic Learning.} 
We then conduct ablations on the semantic learning and DreamBooth~\cite{ruiz2023dreambooth} in our disentangled semantic learning stage. 
Results in Fig.~\ref{fig-Training_Phase_Ablation} demonstrate that omitting either step leads to a degradation in reconstruction fidelity and editing quality. When only the DreamBooth~\cite{ruiz2023dreambooth} is retained, the fine-tuned model fails to accurately reconstruct instances owing to the absence of semantic initialization. Conversely, when only the semantic learning step is preserved, the reconstructed instances exhibit reduced fidelity to the reference image.
\begin{figure}[t]
    \centering
    \includegraphics[width=1\linewidth]{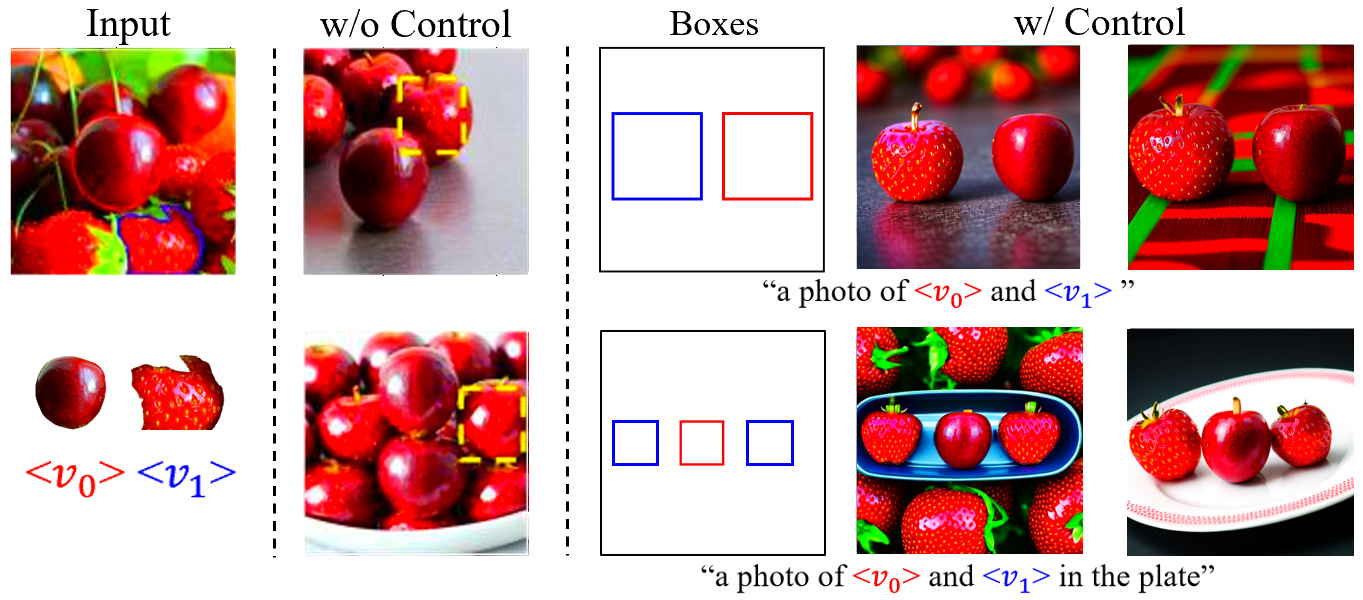}
    \caption{
        \textbf{Qualitative results of precise synthesis with visually similar objects.}
        Semantic leakage are marked with yellow dashed boxes. 
    }
    \label{fig-mult_obj_synthesis_result_show_visual}
\end{figure}

\paragraph{Box Control.}
We conduct ablation studies on Box Control during the precise synthesis stage. As shown in Fig.~\ref{fig-mult_obj_synthesis_result_show_visual}, semantic leakage tends to occur when instances share similar appearances. In contrast, the introduction of Box Control leads to more accurate instance synthesis and reconstruction. 
More qualitative results are presented in Fig.~\ref{fig-mult_obj_synthesis_result_show_semantic} and Fig.~\ref{fig-mult_obj_synthesis_result_show_rare} in Appx.~\ref{appx_subsec-more_qualitative_results}. 

\begin{figure}[htbp]
    \centering
    \includegraphics[width=1\linewidth]{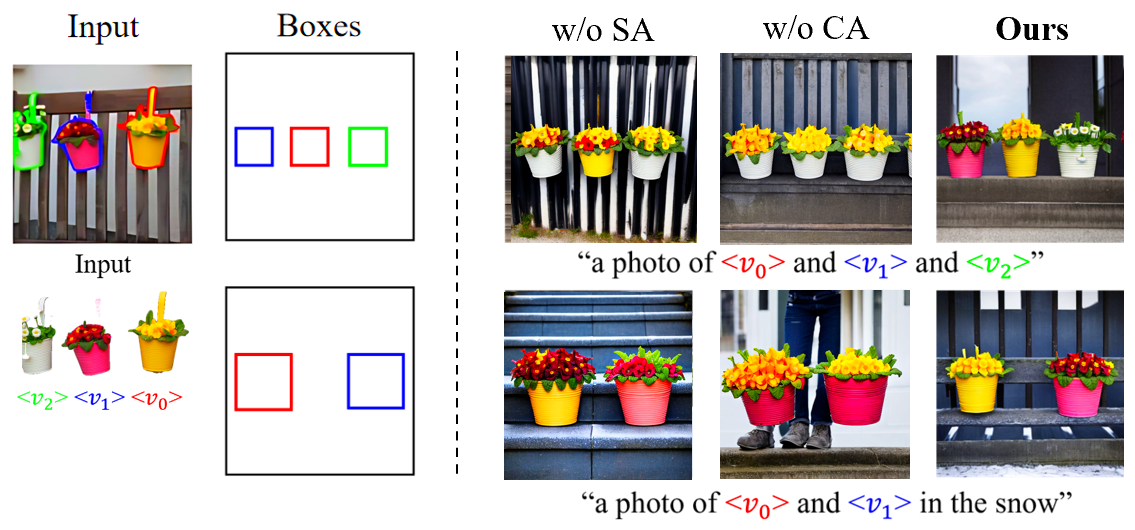}
    \caption{
        \textbf{Ablations on SA/CA control. }
    }
    \label{fig-Attention_Control_Ablation}
\end{figure}

\paragraph{Attention Control.}
We conduct ablations on SA/CA control in the precise synthesis stage, with results presented in Fig.~\ref{fig-Attention_Control_Ablation}. The absence of SA control leads to semantic leakage in two forms. Without in-box control, features from other boxes intrude into a specific box, as exemplified by the appearance of white flowers in the yellow pot (row 1) and red flowers in the yellow pot (row 2). Without out-of-box control, semantics slightly leak into the background, such as the visual features of a pot emerging on the far right (row 1). By contrast, the lack of CA control exerts a more severe impact. It causes pronounced semantic blending between objects, for instance, the yellow pot with red flowers being stacked on top of a red pot (row 2), and it further intensifies semantic leakage in background regions, where visual features of flowers and leaves appear undesirably (row 1). Our method, which incorporates both SA and CA control, effectively mitigates these issues by preventing semantic leakage across instances as well as between boxes and background.


\begin{figure}[htpb]
    \centering
    \includegraphics[width=1\linewidth]{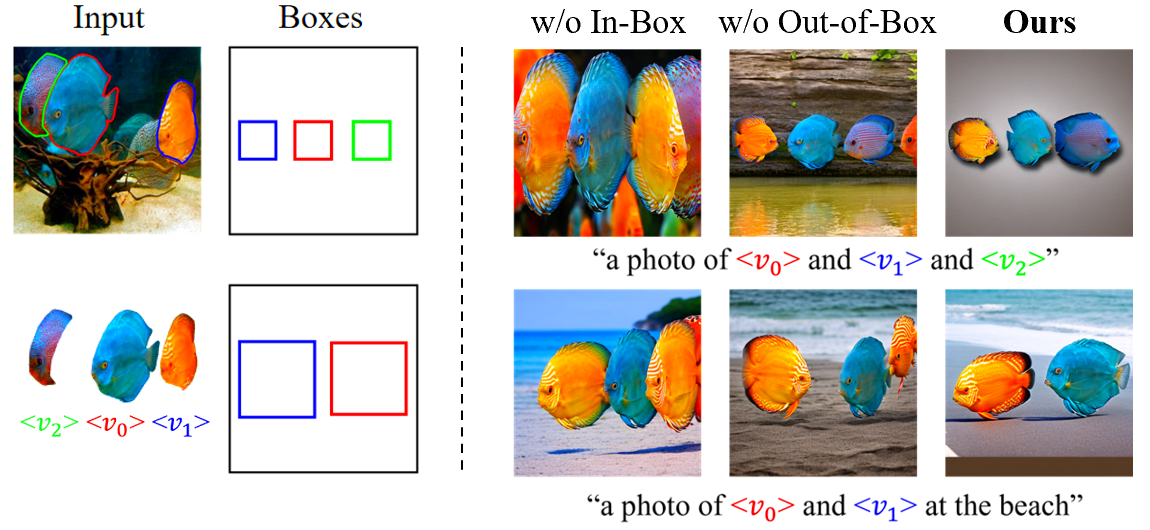}
    \caption{
        \textbf{Ablations on In-Box \& Out-of-Box control.}
    }
    \label{fig-box_control_compare}
\end{figure}
\paragraph{In-Box vs. Out-of-Box Control.}
We conduct ablations on in-box and out-of-box control in the precise synthesis stage, and the results in Fig.~\ref{fig-box_control_compare} highlight their importance. In the absence of in-box control, when computing CA, pixels within the designated box may attend to text features irrelevant to the target instance, resulting in semantic leakage. Similarly, without out-of-box control, the target semantics tend to spill into background regions; for example, the background on the far right exhibits visual features of $\langle v_1 \rangle$. By contrast, our method effectively confines the instance semantics within their corresponding boxes, thereby preventing such leakage.

\begin{figure}[t]
    \centering
    \includegraphics[width=1\linewidth]{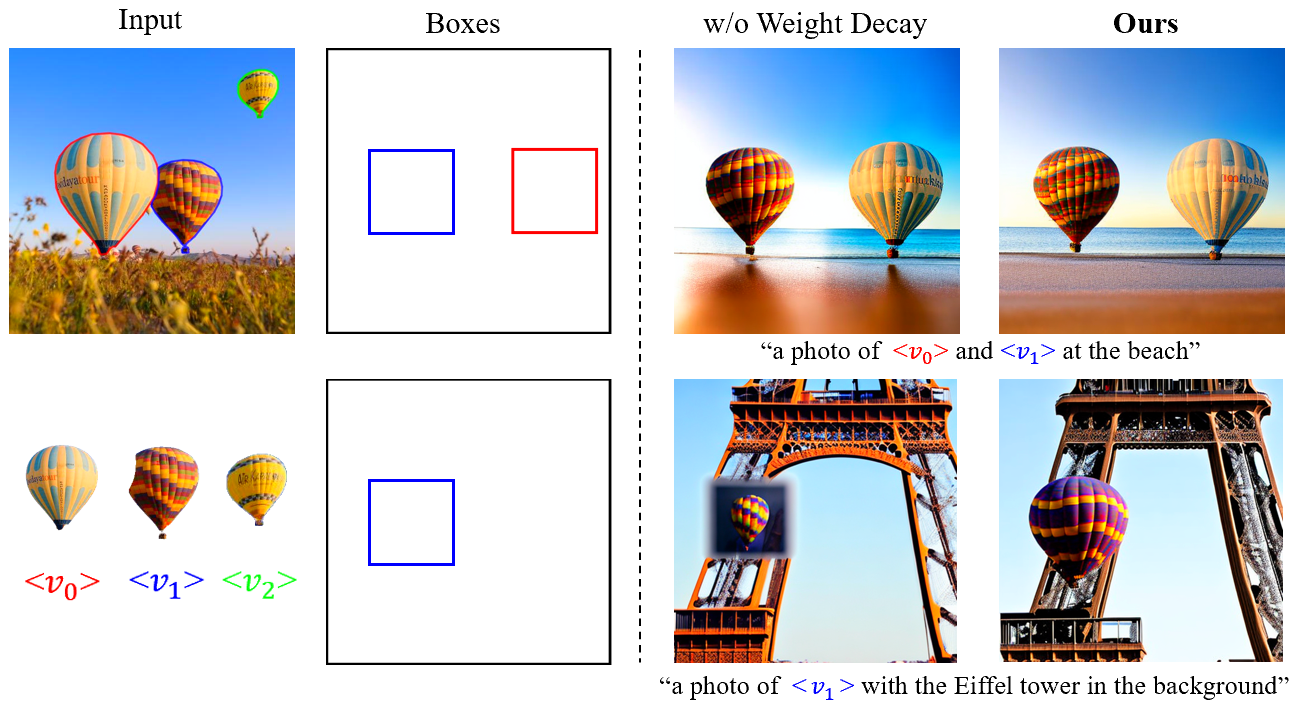}
    \caption{
        \textbf{Ablations on weight decay.}
    }
    \label{fig-without_weight_decay}
\end{figure}

\paragraph{Weight Decay.}
We conduct ablation studies on the effect of weight decay during the precise synthesis stage, with the results presented in Fig.~\ref{fig-without_weight_decay}. In the absence of weight decay (Eq.~\ref{eq-alpha_decay}), the coherence between instances and their surrounding background deteriorates. As shown in Fig.~\ref{fig-without_weight_decay}, the lighting and shadows between the instance and the background become inconsistent (Row 1), and in more severe cases, the instance fails to blend into the background entirely (Row 2).
By contrast, incorporating weight decay promotes smoother semantic transitions and enhances overall visual coherence.



\begin{figure}[t]
    \centering
    \includegraphics[width=1\linewidth]{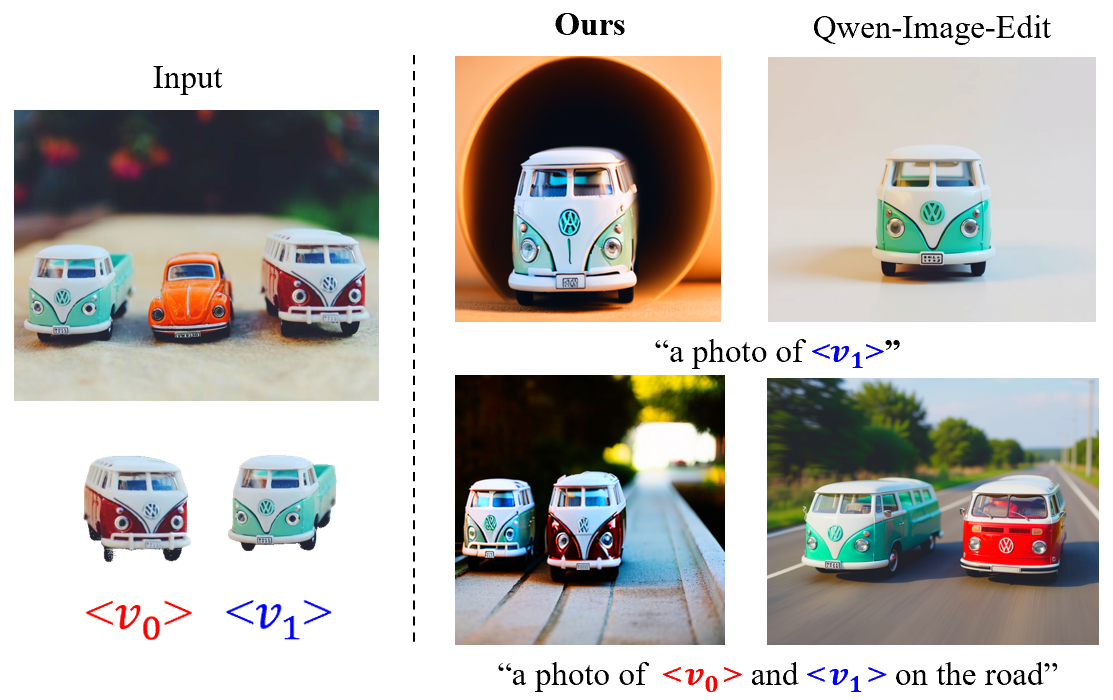}
    \caption{
        \textbf{Comparison with Qwen-image-edit.}
    }
    \label{fig-compare_qwen}
\end{figure}

\subsection{Compare with MLLMs}
To provide further context for our method's performance, we conducted a comparison with the  Multimodal Large Language Model (MLLM), Qwen-Image-Edit~\cite{wu2025qwen}. The quantitative results are presented in Fig.~\ref{fig-compare_qwen}. It should be emphasized that while leading MLLMs typically accept image and text prompt as inputs, they are generally unable to input structured annotations such as masks or bounding boxes. To avoid potential bias introduced by external spatial constraints, all generation results from Qwen-Image-Edit in this study are produced under a free-generation setting—i.e., without any prior bounding box constraints. This design ensures a fair and consistent comparison, enabling a more objective assessment of the proposed method’s effectiveness in instance-level reconstruction.

Results in Fig.~\ref{fig-compare_qwen} demonstrate that although Qwen-Image-Edit also demonstrates significant capability in the context of single-example multi-instance semantic learning, it fails to accurately reconstruct the spatial positioning information of each individual instance in scenarios involving multiple instances. This outcome further serves to demonstrate the efficacy of our proposed methodology.

\section{Conclusion}

We have presented a novel framework for learning and synthesizing multiple instance semantics from a single real-world image. 
During the semantic learning stage, we propose a reward- and penalty-based optimization to disentangle semantics in a coarse-to-fine manner. 
During the synthesis stage, we introduce box control in attention layers to mitigate semantic leakage. 
Our method achieves high-quality and reasonable multi-instance semantic learning and synthesis, excellently balancing editability and instance-consistency. 
It remains robust when dealing with semantically or visually similar instances or rare-seen objects. 
Overall, it provides a practical and generalizable solution for personalized content creation, object-level editing, and controllable multi-object scene reconstruction.








\section*{Acknowledgment}
This work was supported in parts by National Key R\&D Program of China (2024YFB3908500, 2024YFB3908502, 2024YFB3908505), Guangdong Basic and Applied Basic Research Foundation (2023B1515120026), DEGP Innovation Team (2022KCXTD025), SZU Teaching Reform Key Program (JG2024018), and Scientific Development Funds from Shenzhen University.


{\small
\bibliographystyle{cvm}
\bibliography{cvmbib}

\begin{thebibliography}{10}\itemsep=-1pt

\bibitem{avrahami2023break}
O.~Avrahami, K.~Aberman, O.~Fried, D.~Cohen-Or, and D.~Lischinski.
\newblock Break-a-scene: Extracting multiple concepts from a single image.
\newblock In {\em SIGGRAPH Asia 2023 Conference Papers}, pages 1--12, 2023.

\bibitem{bar2023multidiffusion}
O.~Bar-Tal, L.~Yariv, Y.~Lipman, and T.~Dekel.
\newblock Multidiffusion: Fusing diffusion paths for controlled image generation.
\newblock 2023.

\bibitem{caron2021emerging}
M.~Caron, H.~Touvron, I.~Misra, H.~J{\'e}gou, J.~Mairal, P.~Bojanowski, and A.~Joulin.
\newblock Emerging properties in self-supervised vision transformers.
\newblock In {\em Proceedings of the IEEE/CVF international conference on computer vision}, pages 9650--9660, 2021.

\bibitem{chefer2023attend}
H.~Chefer, Y.~Alaluf, Y.~Vinker, L.~Wolf, and D.~Cohen-Or.
\newblock Attend-and-excite: Attention-based semantic guidance for text-to-image diffusion models.
\newblock {\em ACM transactions on Graphics (TOG)}, 42(4):1--10, 2023.

\bibitem{chen2023subject}
W.~Chen, H.~Hu, Y.~Li, N.~Ruiz, X.~Jia, M.-W. Chang, and W.~W. Cohen.
\newblock Subject-driven text-to-image generation via apprenticeship learning.
\newblock {\em Advances in Neural Information Processing Systems}, 36:30286--30305, 2023.

\bibitem{dahary2025decisive}
O.~Dahary, Y.~Cohen, O.~Patashnik, K.~Aberman, and D.~Cohen-Or.
\newblock Be decisive: Noise-induced layouts for multi-subject generation.
\newblock In {\em Proceedings of the Special Interest Group on Computer Graphics and Interactive Techniques Conference Conference Papers}, pages 1--12, 2025.

\bibitem{dahary2024yourself}
O.~Dahary, O.~Patashnik, K.~Aberman, and D.~Cohen-Or.
\newblock Be yourself: Bounded attention for multi-subject text-to-image generation.
\newblock In {\em European Conference on Computer Vision}, pages 432--448. Springer, 2024.

\bibitem{feng2022training}
W.~Feng, X.~He, T.-J. Fu, V.~Jampani, A.~Akula, P.~Narayana, S.~Basu, X.~E. Wang, and W.~Y. Wang.
\newblock Training-free structured diffusion guidance for compositional text-to-image synthesis.
\newblock {\em arXiv preprint arXiv:2212.05032}, 2022.

\bibitem{gal2022image}
R.~Gal, Y.~Alaluf, Y.~Atzmon, O.~Patashnik, A.~H. Bermano, G.~Chechik, and D.~Cohen-Or.
\newblock An image is worth one word: Personalizing text-to-image generation using textual inversion.
\newblock {\em arXiv preprint arXiv:2208.01618}, 2022.

\bibitem{garibi2025tokenverse}
D.~Garibi, S.~Yadin, R.~Paiss, O.~Tov, S.~Zada, A.~Ephrat, T.~Michaeli, I.~Mosseri, and T.~Dekel.
\newblock Tokenverse: Versatile multi-concept personalization in token modulation space.
\newblock {\em arXiv preprint arXiv:2501.12224}, 2025.

\bibitem{ho2020denoising}
J.~Ho, A.~Jain, and P.~Abbeel.
\newblock Denoising diffusion probabilistic models.
\newblock {\em Advances in neural information processing systems}, 33:6840--6851, 2020.

\bibitem{kirillov2023segment}
A.~Kirillov, E.~Mintun, N.~Ravi, H.~Mao, C.~Rolland, L.~Gustafson, T.~Xiao, S.~Whitehead, A.~C. Berg, W.-Y. Lo, et~al.
\newblock Segment anything.
\newblock In {\em Proceedings of the IEEE/CVF international conference on computer vision}, pages 4015--4026, 2023.

\bibitem{kumari2023multi}
N.~Kumari, B.~Zhang, R.~Zhang, E.~Shechtman, and J.-Y. Zhu.
\newblock Multi-concept customization of text-to-image diffusion.
\newblock In {\em Proceedings of the IEEE/CVF conference on computer vision and pattern recognition}, pages 1931--1941, 2023.

\bibitem{li2023gligen}
Y.~Li, H.~Liu, Q.~Wu, F.~Mu, J.~Yang, J.~Gao, C.~Li, and Y.~J. Lee.
\newblock Gligen: Open-set grounded text-to-image generation.
\newblock In {\em Proceedings of the IEEE/CVF conference on computer vision and pattern recognition}, pages 22511--22521, 2023.

\bibitem{lin2014microsoft}
T.-Y. Lin, M.~Maire, S.~Belongie, J.~Hays, P.~Perona, D.~Ramanan, P.~Doll{\'a}r, and C.~L. Zitnick.
\newblock Microsoft coco: Common objects in context.
\newblock In {\em Computer vision--ECCV 2014: 13th European conference, zurich, Switzerland, September 6-12, 2014, proceedings, part v 13}, pages 740--755. Springer, 2014.

\bibitem{peebles2023scalable}
W.~Peebles and S.~Xie.
\newblock Scalable diffusion models with transformers.
\newblock In {\em Proceedings of the IEEE/CVF international conference on computer vision}, pages 4195--4205, 2023.

\bibitem{phung2024grounded}
Q.~Phung, S.~Ge, and J.-B. Huang.
\newblock Grounded text-to-image synthesis with attention refocusing.
\newblock In {\em Proceedings of the IEEE/CVF Conference on Computer Vision and Pattern Recognition}, pages 7932--7942, 2024.

\bibitem{ramesh2022hierarchical}
A.~Ramesh, P.~Dhariwal, A.~Nichol, C.~Chu, and M.~Chen.
\newblock Hierarchical text-conditional image generation with clip latents.
\newblock {\em arXiv preprint arXiv:2204.06125}, 1(2):3, 2022.

\bibitem{rombach2022high}
R.~Rombach, A.~Blattmann, D.~Lorenz, P.~Esser, and B.~Ommer.
\newblock High-resolution image synthesis with latent diffusion models.
\newblock In {\em Proceedings of the IEEE/CVF conference on computer vision and pattern recognition}, pages 10684--10695, 2022.

\bibitem{ruiz2023dreambooth}
N.~Ruiz, Y.~Li, V.~Jampani, Y.~Pritch, M.~Rubinstein, and K.~Aberman.
\newblock Dreambooth: Fine tuning text-to-image diffusion models for subject-driven generation.
\newblock In {\em Proceedings of the IEEE/CVF conference on computer vision and pattern recognition}, pages 22500--22510, 2023.

\bibitem{song2020denoising}
J.~Song, C.~Meng, and S.~Ermon.
\newblock Denoising diffusion implicit models.
\newblock {\em arXiv preprint arXiv:2010.02502}, 2020.

\bibitem{tewel2023key}
Y.~Tewel, R.~Gal, G.~Chechik, and Y.~Atzmon.
\newblock Key-locked rank one editing for text-to-image personalization.
\newblock In {\em ACM SIGGRAPH 2023 conference proceedings}, pages 1--11, 2023.

\bibitem{tunanyan2023multi}
H.~Tunanyan, D.~Xu, S.~Navasardyan, Z.~Wang, and H.~Shi.
\newblock Multi-concept t2i-zero: Tweaking only the text embeddings and nothing else.
\newblock {\em arXiv preprint arXiv:2310.07419}, 2023.

\bibitem{vaswani2017attention}
A.~Vaswani, N.~Shazeer, N.~Parmar, J.~Uszkoreit, L.~Jones, A.~N. Gomez, {\L}.~Kaiser, and I.~Polosukhin.
\newblock Attention is all you need.
\newblock {\em Advances in neural information processing systems}, 30, 2017.

\bibitem{voynov2023p+}
A.~Voynov, Q.~Chu, D.~Cohen-Or, and K.~Aberman.
\newblock p+: Extended textual conditioning in text-to-image generation.
\newblock {\em arXiv preprint arXiv:2303.09522}, 2023.

\bibitem{wei2023elite}
Y.~Wei, Y.~Zhang, Z.~Ji, J.~Bai, L.~Zhang, and W.~Zuo.
\newblock Elite: Encoding visual concepts into textual embeddings for customized text-to-image generation.
\newblock In {\em Proceedings of the IEEE/CVF International Conference on Computer Vision}, pages 15943--15953, 2023.

\bibitem{wu2025qwen}
C.~Wu, J.~Li, J.~Zhou, J.~Lin, K.~Gao, K.~Yan, S.-m. Yin, S.~Bai, X.~Xu, Y.~Chen, et~al.
\newblock Qwen-image technical report.
\newblock {\em arXiv preprint arXiv:2508.02324}, 2025.

\bibitem{xie2023boxdiff}
J.~Xie, Y.~Li, Y.~Huang, H.~Liu, W.~Zhang, Y.~Zheng, and M.~Z. Shou.
\newblock Boxdiff: Text-to-image synthesis with training-free box-constrained diffusion.
\newblock In {\em Proceedings of the IEEE/CVF International Conference on Computer Vision}, pages 7452--7461, 2023.

\bibitem{yang2023reco}
Z.~Yang, J.~Wang, Z.~Gan, L.~Li, K.~Lin, C.~Wu, N.~Duan, Z.~Liu, C.~Liu, M.~Zeng, et~al.
\newblock Reco: Region-controlled text-to-image generation.
\newblock In {\em Proceedings of the IEEE/CVF Conference on Computer Vision and Pattern Recognition}, pages 14246--14255, 2023.

\bibitem{zheng2023layoutdiffusion}
G.~Zheng, X.~Zhou, X.~Li, Z.~Qi, Y.~Shan, and X.~Li.
\newblock Layoutdiffusion: Controllable diffusion model for layout-to-image generation.
\newblock In {\em Proceedings of the IEEE/CVF Conference on Computer Vision and Pattern Recognition}, pages 22490--22499, 2023.

\end{thebibliography}
}

\newpage
\appendix


\newpage

\appendix

\newcommand{\secondtitle}[1]{
    {
        \centering
        \normalfont
        \Large\textbf{#1}
        \vspace{1\baselineskip}
    }
}

\secondtitle{
    {
        Appendix
    }
}

\section{More Discussion on Semantic Leakage}
\label{appx_sec-more_discussion_on_semantic_leakage}

\subsection{More attention query results}

\begin{figure}[H]
    \centering
    \includegraphics[width=1\linewidth]{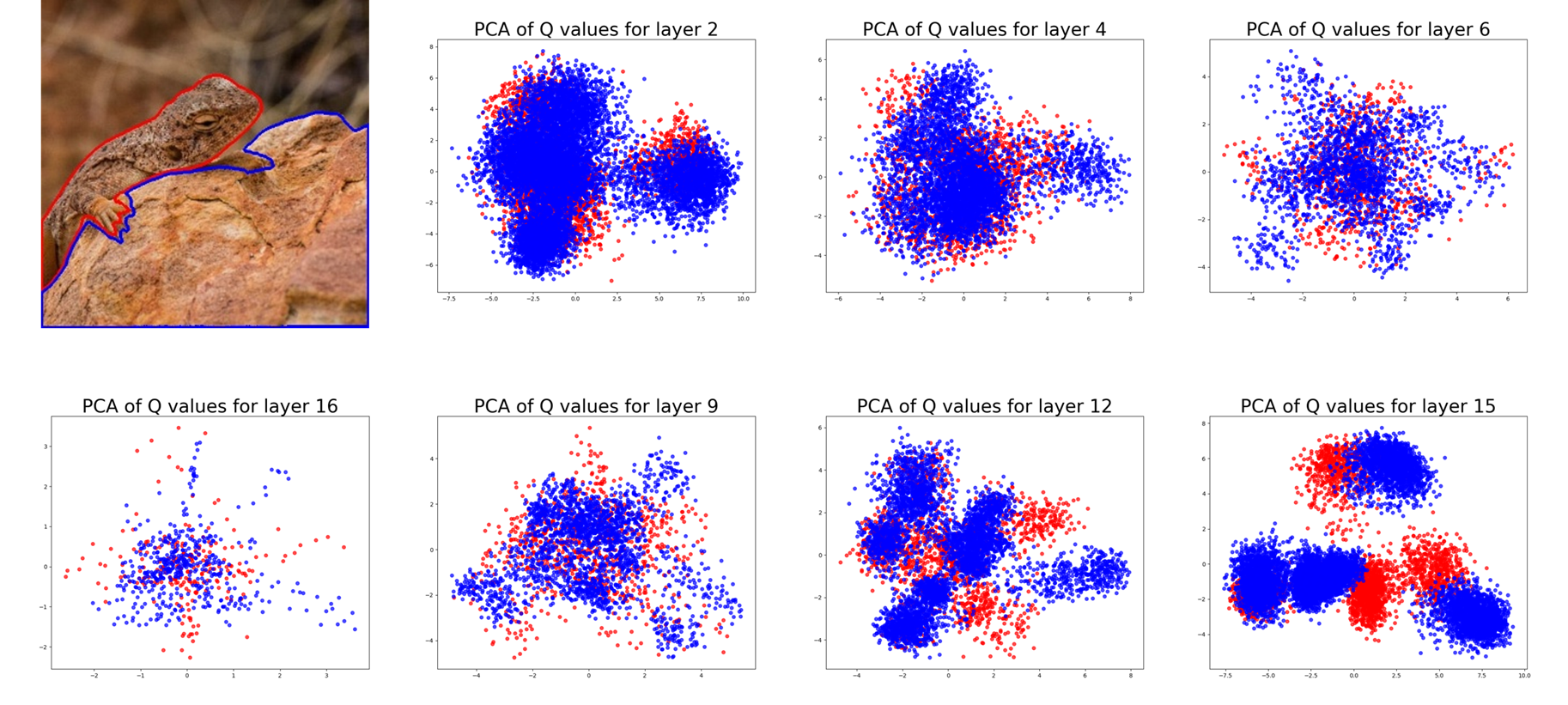}
    \caption{
        \textbf{More visualization results of the Query in different layers of attention features.}
    }
    \label{fig-more_PCA_query_result}
\end{figure}

To further elucidate the underlying causes of information leakage, we visualize the query features across multiple layers, as shown in Fig.~\ref{fig-more_PCA_query_result}. It can be observed that semantic entanglement is not confined to specific layers but instead manifests consistently throughout the hierarchical structure. This widespread entanglement ultimately gives rise to confusion and information leakage during the synthesis process.

\subsection{Non-directional Convergence States of Reward-based Mechanisms}

As discussed in Sec.~\ref{semantic_learning} in the main paper, the reward-based mechanisms only encourage alignment rather than discouraging misalignment. 

\begin{figure}[htbp]
    \centering
    \includegraphics[width=1\linewidth]{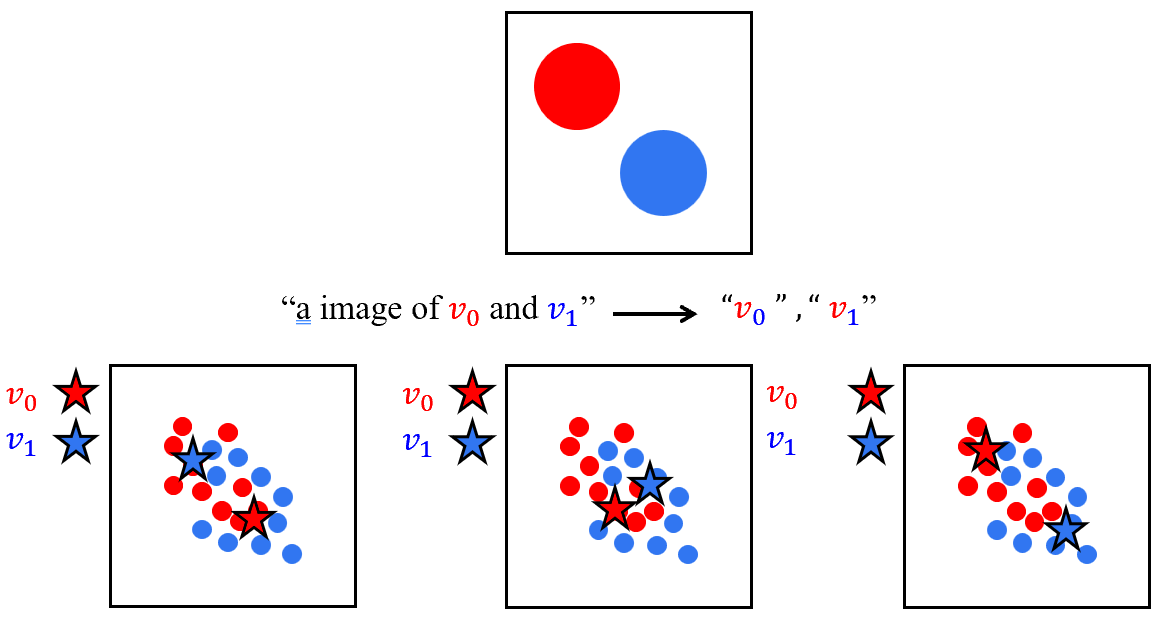}
    \caption{
        \textbf{Possible convergence states of reward-based mechanisms. }
    }
    \label{fig-without_control_vec_train_sample}
\end{figure}

Fig.~\ref{fig-without_control_vec_train_sample} presents three possible convergence states when solely employing reward-based mechanisms in semantic entanglement scenarios. In the first case (from left to right), the two text embeddings mutually learn each other’s target semantics, a situation that may arise because this state also minimizes $\mathcal{L}_{\text{CA}}^{\text{reward}}$. In the second case, the embeddings fail to achieve complete disentanglement and eventually stabilize at the boundary between the two target semantics. The third case corresponds to the desired outcome, where the embeddings converge to the correct disentangled representations. These convergence behaviors underscore the inherent non-directional nature of reward-based mechanisms, which proves insufficient for precise semantic learning and synthesis.


\subsection{Attention Visualization}
\label{appx_subsec-attention_visualization}

Semantically correlated tokens share similar key features $K$ in CA layers, which leads to ambiguity in how the image query features $Q$ respond to the text embeddings during attention computation. 
When a pixel's query features $Q$ are simultaneously similar to the key features $K$s from multiple tokens, the derived attention weights will be distributed across those tokens with comparable magnitude. 
Then, this pixel aggregates the $K$s from multiple tokens when deriving the value features $V$ with weighted average, thus blending visual features from different semantics. 


\begin{figure}[htbp]
    \centering
    \includegraphics[width=1\linewidth]{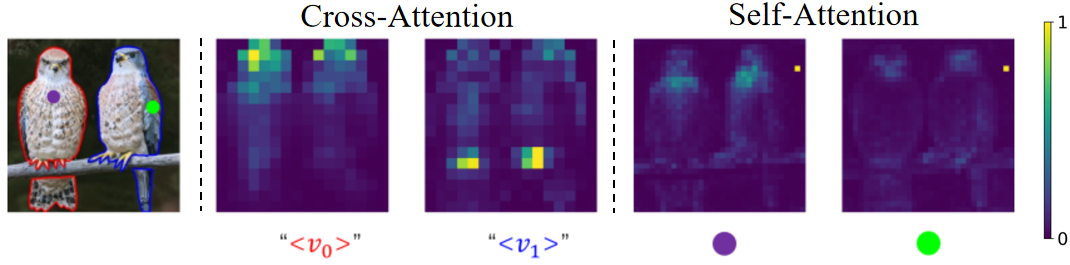}
    \caption{
        \textbf{Visualization of semantic leakage in attention layers. }
        CA weights for tokens $\langle v_0 \rangle$ (red silhouette) and $\langle v_1 \rangle$ (blue silhouette) are displayed, while SA weights for the two pixels marked with purple and green dots are visualized. 
    }
    \label{fig-leakage_attn}
\end{figure}
To analyze the occurrence of semantic leakage, we visualize the attention weights of two tokens (derived from the disentangled semantic learning stage) and two pixels at timestep $t=500$. As shown in Fig.~\ref{fig-leakage_attn}, semantically associated tokens receive elevated attention weights in both target regions, indicating semantic leakage in CA layers. Similarly, pixels corresponding to single-instance semantics exhibit high attention weights in the pixel regions of another target, revealing semantic leakage in SA layers. These observations demonstrate that leakage can arise in both SA and CA layers, which motivates our introduction of box control in both types of attention layers during the precise synthesis stage (Sec.~\ref{synthesis_control} in the main paper).



\section{Semantic Learning Details}
\label{appx_sec-semantic_learning_details}

\subsection{Pseudo-code}
\begin{figure}[t]
\centering
\resizebox{0.9\linewidth}{!}{%
\begin{minipage}{\linewidth}
\begin{algorithm}[H]
\caption{Disentangled Semantic Learning}
\label{alg-semantic_learning}
\begin{algorithmic}[1]
\Require Image $I$, masks $\{M_i\}_{i=0}^{N-1}$, placeholders $\{\langle v_i\rangle\}_{i=0}^{N-1}$, 
\Statex \quad CA layers $\mathcal{L}_{\mathrm{CA}}$, iterations $E$, coarse cutoff $e_{\mathrm{coarse}}$, $E_{\mathrm{stage1}}$
\Ensure Embeddings $\{\mathbf{v}_{\mathrm{ec}}^i\}_{i=0}^{N-1}$
\State Initialize $\{\mathbf{v}_{\mathrm{ec}}^i\}$ from CLIP; freeze other tokens
\For{$e=1$  \textbf{to}  $E$}
  \If{$e\le E_{\mathrm{stage1}}$} 
    \State Sample $\Lambda\subseteq\{0,\dots,N-1\}$; $M_{\mathrm{rec}}\leftarrow\bigcup_{i\in\Lambda}M_i$
    \State Sample $t$, obtain $z_t$, predict $\hat{\epsilon}=\epsilon_\phi(z_t,t,c_\theta)$
    \State $\mathcal{L}_{\mathrm{rec}}\leftarrow \mathbb{E}\|M_{\mathrm{rec}}\odot(\epsilon-\hat{\epsilon})\|_2^2$
    \State $\mathcal{L}_{\mathrm{attn}}\leftarrow 0$
    \ForAll{$j\in\Lambda$}\ForAll{$l\in\mathcal{L}_{\mathrm{CA}}$}
      \If{$e<e_{\mathrm{coarse}}$}
        \State $\mathcal{L}_{\mathrm{attn}}\!+\!=\!\|\alpha M_j^l-A_j^l\|_2^2$
      \Else
        \State $\mathcal{L}_{\mathrm{attn}}\!+\!=\!\|(1-M_j^l)\odot A_j^l\|_2^2$
      \EndIf
    \EndFor\EndFor
    \State $\mathcal{L}\leftarrow\lambda_{\mathrm{rec}}\mathcal{L}_{\mathrm{rec}}+\lambda_{\mathrm{attn}}\mathcal{L}_{\mathrm{attn}}$
    \State Update $\{\mathbf{v}_{\mathrm{ec}}^i\}$ w.r.t. $\mathcal{L}$
  \Else 
    \State Compute DreamBooth loss $\mathcal{L}_{\mathrm{DB}}$
    \State Jointly update $\{\mathbf{v}_{\mathrm{ec}}^i\}$, U-Net, text-encoder w.r.t. $\mathcal{L}_{\mathrm{DB}}$
  \EndIf
\EndFor
\State \Return $\{\mathbf{v}_{\mathrm{ec}}^i\}$, U-Net, text-encoder
\end{algorithmic}
\end{algorithm}
\end{minipage}
}
\end{figure}

\begin{figure}[t]
\centering
\resizebox{0.9\linewidth}{!}{%
\begin{minipage}{\linewidth}
\begin{algorithm}[H]
\caption{Precise Synthesis}
\label{alg-semantic_synthesis}
\begin{algorithmic}[1]
\Require Embeddings $\{\mathbf{v}_{\mathrm{ec}}^i\}$, initial boxes $\{B_i\}$, latent $z_T$, bound stage $T_{\mathrm{bound}}$, step size $\beta$, update interval $k$
\Ensure Synthesized image $I_{\mathrm{syn}}$
\For{$t=T$ \textbf{downto} $1$}
  \State Compute CA and SA attention $\{A_l\}_{l\in\mathcal{L}}$
  \State Apply mask control using current masks $\{M_i^t\}$ (in-box permitted, out-of-box suppressed)
  \If{$t \le T_{\mathrm{bound}}$} 
    \State $\mathcal{L}_{\mathrm{reward}}\leftarrow\sum_{i,l}\mathcal{R}(A_l,M_i^t)$
    \State $\mathcal{L}_{\mathrm{penalty}}\leftarrow\sum_{i,l}\mathcal{P}(A_l,M_i^t)$
    \State $\mathcal{L}\leftarrow\mathcal{L}_{\mathrm{reward}}+\alpha(t)\mathcal{L}_{\mathrm{penalty}}$
    \State $z_t\leftarrow z_t-\beta\nabla_{z_t}\mathcal{L}$
  \EndIf
  \If{$t > T_{\mathrm{bound}}$ \textbf{and} $t \bmod k = 0$} 
    \State $M_i^{cross} \leftarrow \text{ComputeCAMasks}(CA_{maps}, s, \sigma_{noun})$
    \State $C_j^{self} \leftarrow \text{KMeans}(SA_{features}, prev_{centers})$
    \State $M_i^{l} \leftarrow \text{Assign}(M_i^{cross}, C_j^{self}, \sigma_{cluster})$
    \State Update masks $\{M_i^t\}$ by K-means clustering on self-attention features
  \EndIf
  \State $z_{t-1}\leftarrow\mathrm{DDIM\_Step}(z_t,\epsilon_\phi,t)$
\EndFor
\State \Return $\mathrm{VAE\_Decode}(z_0)$
\end{algorithmic}
\end{algorithm}
\end{minipage}
}
\end{figure}

\subsection{Timestep Selection}





\begin{figure}[htbp]
    \centering
    \includegraphics[width=1\linewidth]{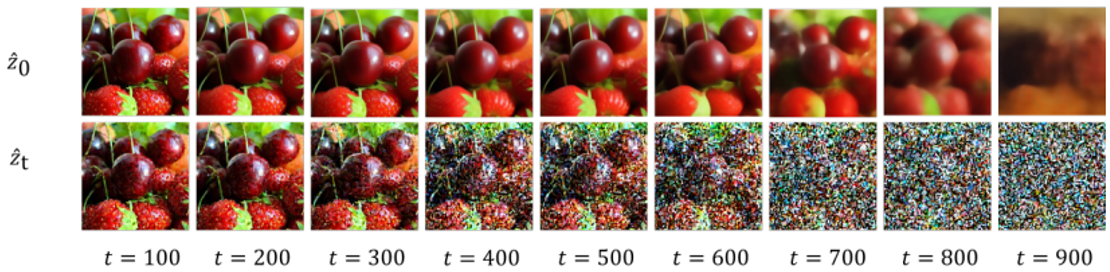}
    \caption{
        $\hat{z}_0$ \textbf{reconstructed with predicted noise residuals at different timesteps. }
    }
    \label{fig-DDIM_inference_and_inversion}
\end{figure}

To achieve accurate semantic learning, we empirically select timesteps for attention loss computation based on the following observations. 

Specifically, we add noise to $z_0$ via deterministic DDIM~\cite{song2020denoising}, and reconstruct $\hat{z}_0$ with the predicted noise residual $\epsilon_\phi (z_t, t, c_\theta)$. 
Results in Fig.~\ref{fig-DDIM_inference_and_inversion} shows that $\hat{z}_0$ exhibits varying degrees of changes in appearance and structural layout when $t \geq 700$. 


\begin{figure}[htbp]
    \centering
    \includegraphics[width=1\linewidth]{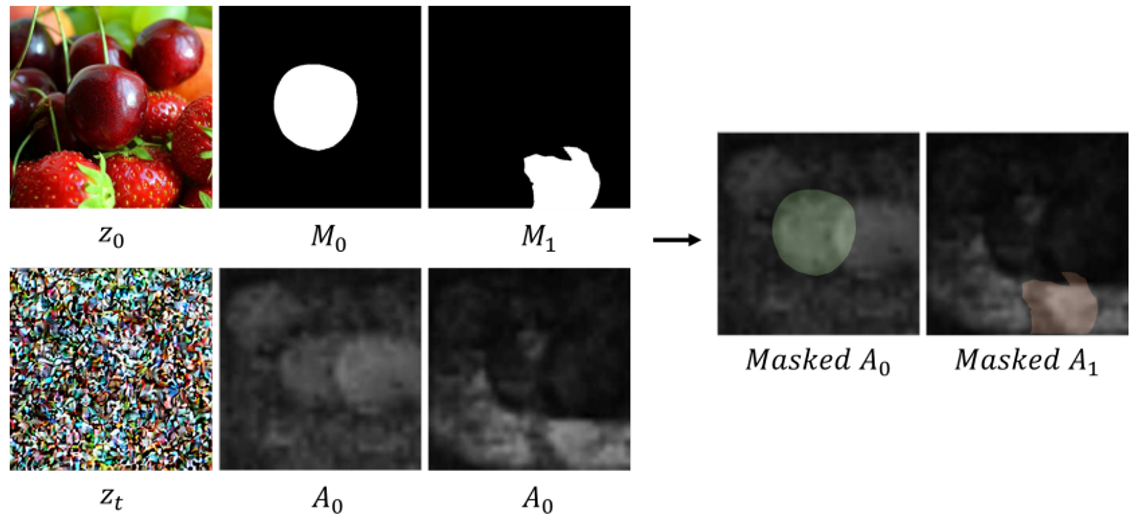}
    \caption{
        \textbf{Illustration of semantic confusion at high noise level. }
    }
    \label{fig-add_noise_cause_change}
\end{figure}

To further investigate the impact of high noise levels on CA control, we visualize the attention weights at timestep $t = 800$ in Fig.~\ref{fig-add_noise_cause_change}. 
Since the input masks $M_i$ ($i = 0, 1$) cannot adaptively adjust their spatial layout in response to the layout changes of $z_t$ induced by adding noise, directly applying $M_i$ for attention control results in layout mismatch, leading to semantic confusion during the semantic learning stage. 


Therefore, we restrict the computation of attention loss to timesteps $t \leq 700$. 
This strategy does not cause performance degradation due to the lack of supervision at $t > 700$, because text conditions are only injected to CA layers in Stable Diffusion U-Net~\cite{rombach2022high} and it's independent of timestep $t$, which produces identical key and value features in any CA layers regardless of $t$. 


\subsection{Attention Layer Selection}

\begin{figure}[H]
    \centering
    \includegraphics[width=1\linewidth]{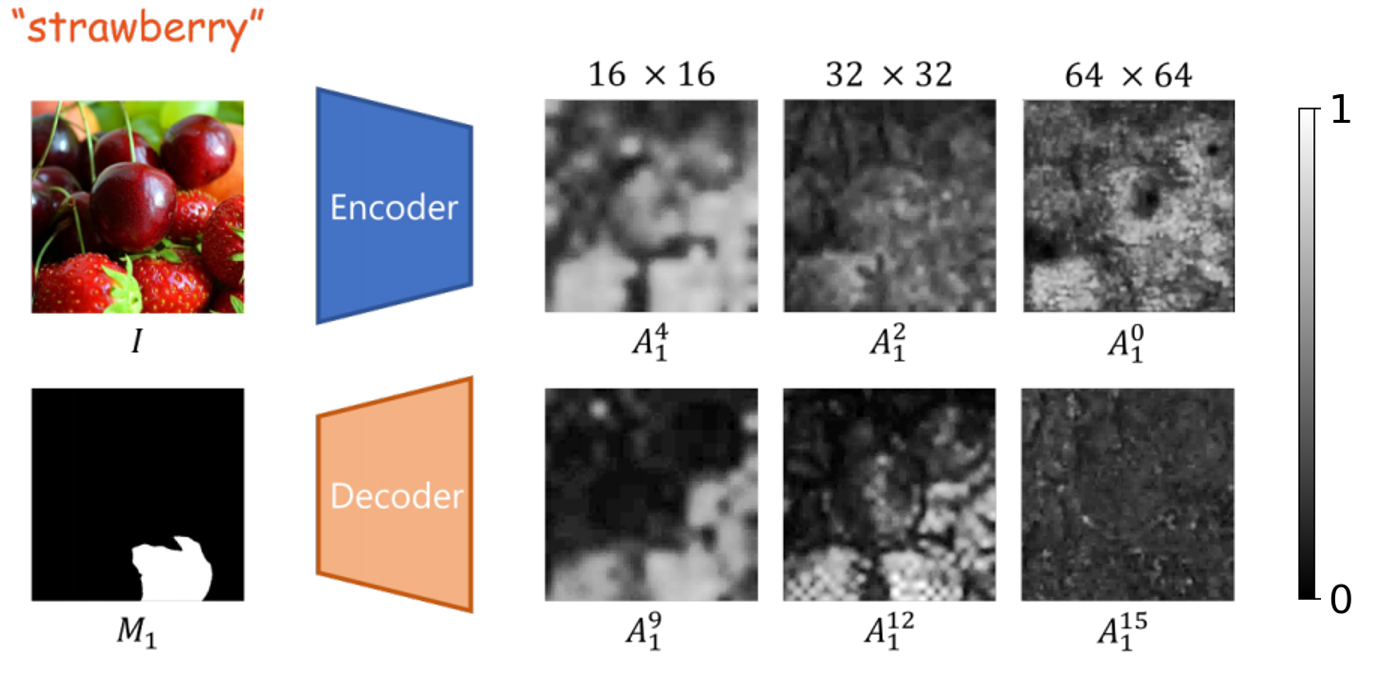}
    \caption{
        \textbf{Visualization of attention matrices of the token ``\texttt{strawberry}" in various CA layers. }
    }
    \label{fig-difference_CA_label}
\end{figure}

Different CA layers within the U-Net focus on distinct levels of information~\cite{voynov2023p+}, which can be summarized as: shallower layers encode appearance and style semantics, while deeper ones encode structural and categorical semantics. 




To determine which CA layers should be used for attention loss computation, we first visualize the attention matrices of the token ``\texttt{strawberry}" across different CA layers, as shown in Fig.~\ref{fig-difference_CA_label}. The encoder layers display scattered and chaotic attention patterns, which can be attributed to the incomplete integration of image features during the encoding stage. By contrast, the decoder layers produce more concentrated attention distributions: the $16 \times 16$ layers effectively capture categorical semantics, the $32 \times 32$ layers fail to disentangle appearance semantics, and the $64 \times 64$ layers only contribute marginally to feature fine-tuning. Based on these observations, we restrict the computation of attention loss to decoder CA layers, thereby enabling more accurate learning of both categorical and appearance semantics.


\section{Experimental Details}
\label{appx_sec-experimentall_details}

\subsection{Dataset}
\label{appx_subset-dataset}

Our dataset consists of 30 images from COCO dataset~\cite{lin2014microsoft} and Unsplash (\href{https://unsplash.com/}{https://unsplash.com/}). 
Each image contains at least two instances, with each instance occupying at least $15\%$ of the full image. 
Particularly, 15 images contain instances with high semantic or visual similarity, while others include semantically independent ones. 

For pre-processing, we resize the images such that the shorter side is 512 pixels, center-cropped to 512x512, and derive the semantic segmentation with Segment Anything Model (SAM)~\cite{kirillov2023segment}. 


\subsection{Environment}

All experiments are conducted on an NVIDIA Quadro P6000 GPU (24 GB VRAM), and run on a Ubuntu 20.04.3 LTS operating system with an Intel Xeon E5-2650 v4 processor. 

\subsection{Hyper-parameters}

\begin{itemize}
    \item Reward-based mechanism (Eq.~\ref{reward-base_loss_funtion} in the main paper): 
        $\alpha = 0.5$; 
    \item Learning rates for semantic learning (Sec.~\ref{semantic_learning} in the main paper): 
        $5 \times 10^{-3}$ for the first stage, $2 \times 10^{-6}$ for the second stage; 
    \item Loss weights for semantic learning (Eq.~\ref{learning_stage_total_loss} in the main paper): 
        $\lambda_{\text{rec}} = 1$, $\lambda_{\text{attn}} = 0.01$; 
    \item Loss weights for the loss function in precise synthesis(Eq.~\ref{eq-alpha_decay} in the main paper):
    $\alpha_{max}=0.5$, $\alpha_{min}=0.2$, $\alpha_{final}=0.1$, $S_1 =3$, $N=15$
    \item Loss weights for precise synthesis (Eq.~\ref{eq-combined_attn_loss} in the main paper): 
        $\lambda_{\text{attn}} ^{\text{SA}} = 0.5$, $\lambda_{\text{attn}} ^{\text{CA}} = 1.5$. 
\end{itemize}


\subsection{Evaluation Metrics}
\label{appx_subsec-evaluation_metrics}

\paragraph{Metrics}

We adopt the following metrics for quantitative evaluation:

\begin{itemize}
  \item \textbf{SIM-C}: Semantic similarity between the image and text features; 
  \item \textbf{SIM-D}: Semantic similarity between the generated samples and the corresponding image features; 
  \item \textbf{NSIM-D}: Semantic similarity between the generated samples and the irrelevant image features; 
  \item \textbf{HPS v2}: A human preference metric that captures perceptual and semantic fidelity from human evaluations; 
\item \textbf{Composite Score (CS)}: Providing an integrated measure of overall performance.

\end{itemize}

SIM-C is computed as the cosine similarity between text and image embeddings extracted by a pre-trained CLIP model~\cite{ramesh2022hierarchical}. 
SIM-D and NSIM-D are computed as the cosine similarity between the generated samples and, respectively, the corresponding and irrelevant image features extracted by a pre-trained DINO model~\cite{caron2021emerging}. 
The Composite Score (CS) is then defined as 
\[
\text{CS} = \text{SIM-C} \times \text{SIM-D} \times (1 - \text{NSIM-D}),
\]
serving as an integrated measure of overall performance.



\paragraph{Evaluation Dimensions}

In the semantic learning stage, SIM-C evaluates the editability of the learned representations, while SIM-D and NSIM-D jointly assess the instance consistency of reference images from different perspectives. In the precise synthesis stage, HPS v2 and the Human Preference Metric are additionally introduced to measure the perceptual quality of image generation. Moreover, a Composite Score is employed across both stages to evaluate the overall balance among multiple metrics. Together, these metrics provide a comprehensive assessment of semantic learning and synthesis.



\paragraph{User Study}
\label{appx-user_study}

\begin{figure}[htbp]
    \centering
    \includegraphics[width=1\linewidth]{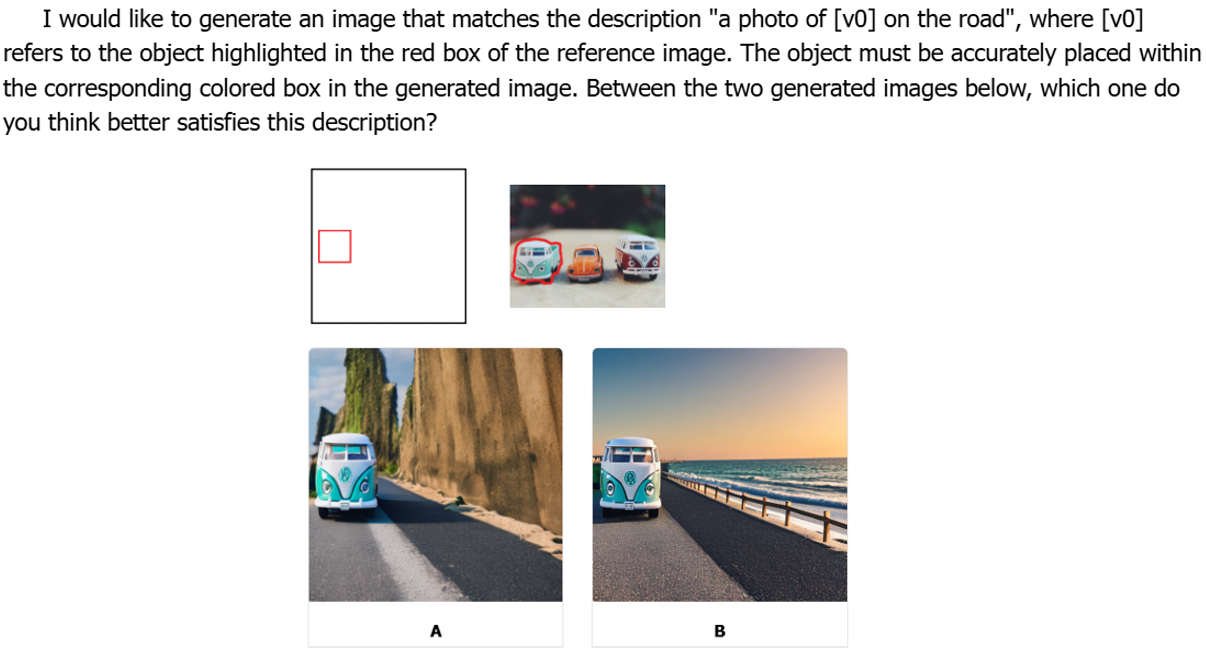}
    \caption{
        \textbf{The example of user study questionnaire.}
    }
    \label{fig-user_study_example}
\end{figure}

To investigate human preferences among various synthesis methods, we conducted a user study via a structured questionnaire. Each question, as illustrated in Fig.~\ref{fig-user_study_example}, presents a pairwise comparison between our method and that of a randomly selected method. To ensure comprehensive and fair evaluation, we systematically cover diverse instances and prompts across all compared methods. The final results were collected and quantified by computing the win rate of our approach over others in each pairwise comparison, serving as the primary metric for the user study.

\subsection{Sampling Strategies}
\label{appx_subsec-sampling_strategies}

\paragraph{Separate Sampling.}

We isolate each instance with a mask, which is subsequently paired with a text prompt in the form of ``\texttt{a photo of} $\langle v_i \rangle$". 
This enforces each text embedding to only interact with its corresponding instance during semantic learning. 

\paragraph{Joint Sampling.} 

In each iteration, we randomly sample $k$ ($1 \leq k \leq n$) instances from the $N$ target instances per iteration, forming an index set $\Lambda = \{i_1, i_2, \dots, i_k\}$, and derive the combined mask as
\begin{equation}
    M_{\text{rec}} 
    = 
    \bigcup_{i \in \Lambda} M_i
\end{equation}
This strategy allows for $(2^N - 1)$ unique mask combinations, providing effective data augmentation. 

\begin{figure}[htbp]
    \centering
    \includegraphics[width=1\linewidth]{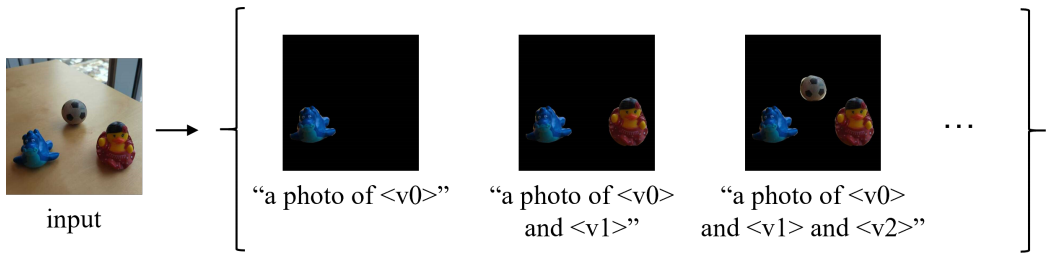}
    \caption{
        \textbf{Illustration of joint sampling.}
        Each mask combination is associated with a distinct prompt.
    }
    \label{fig-ti_and_db_input_process}
\end{figure}

An illustration is provided in Fig.~\ref{fig-ti_and_db_input_process}. 

\subsection{Text Prompts}

We adopt text prompts in the form of ``\texttt{a photo of ...}" for reconstruction experiments, while adopting those in Tab.~\ref{tab-text_prompts} for editability experiments. 

\begin{table}[htbp]
    \centering
    \caption{
        \textbf{Text prompts for editability experiments. }
    }
    \label{tab-text_prompts}
    \begin{tabular}{p{0.425\textwidth}}
        \toprule
        \hspace{2em}\textbf{Prompt Template} \\
        \midrule
        \hspace{2em}"a photo of \dots\ at the beach" \\
        \hspace{2em}"a photo of \dots\ in the jungle" \\
        \hspace{2em}"a photo of \dots\ in the snow" \\
        \hspace{2em}"a photo of \dots\ in the street" \\
        \hspace{2em}"a photo of \dots\ on top of a pink fabric" \\
        \hspace{2em}"a photo of \dots\ floating on top of water" \\
        \hspace{2em}"a photo of \dots\ on top of a wooden floor" \\
        \hspace{2em}"a photo of \dots\ with a city in the background" \\
        \hspace{2em}"a photo of \dots\ with a mountain in the background" \\
        \hspace{2em}"a photo of \dots\ with the Eiffel tower in the background" \\
        \bottomrule
    \end{tabular}
\end{table}

\section{More Experimental Results}
\label{appx_sec-more_experimental_results}




\subsection{More Qualitative Results}
\label{appx_subsec-more_qualitative_results}

\begin{figure}[htbp]
    \centering
    \includegraphics[width=1\linewidth]{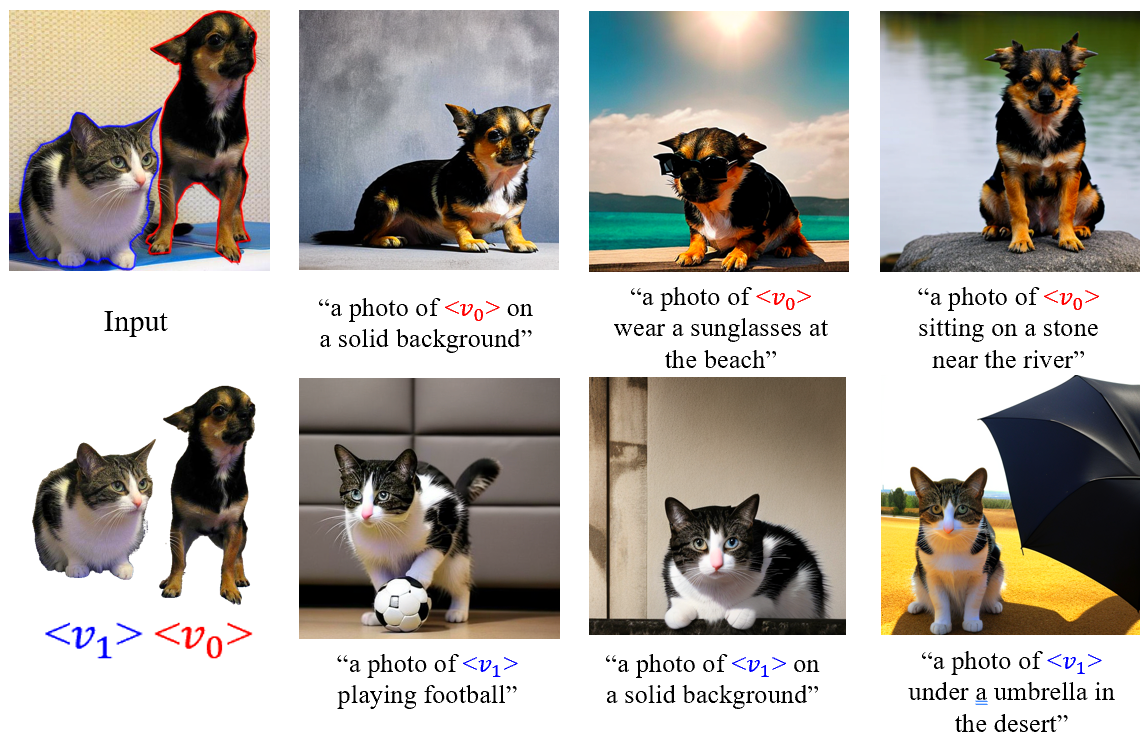}
    \caption{
        \textbf{Qualitative results of semantic learning with semantically similar objects. }
    }
    \label{fig-mult_obj_learning_result_semantic}
\end{figure}

\begin{figure}[htbp]
    \centering
    \includegraphics[width=1\linewidth]{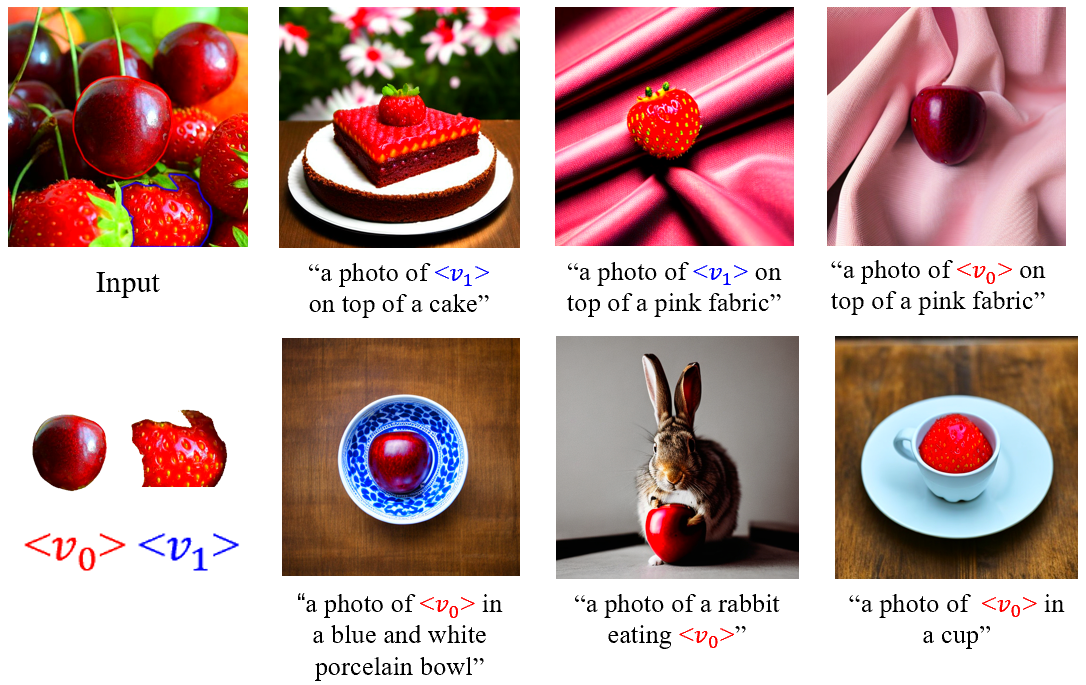}
    \caption{
        \textbf{Qualitative results of semantic learning with visually similar objects. }
    }
    \label{fig-mult_obj_learning_result_visual}
\end{figure}

\paragraph{Semantic Learning.}
More qualitative results, such as images with semantic or visual similarities are shown in Fig.~\ref{fig-mult_obj_learning_result_semantic} and Fig.~\ref{fig-mult_obj_learning_result_visual}, respectively. 

\begin{figure}[htbp]
    \centering
    \includegraphics[width=1\linewidth]{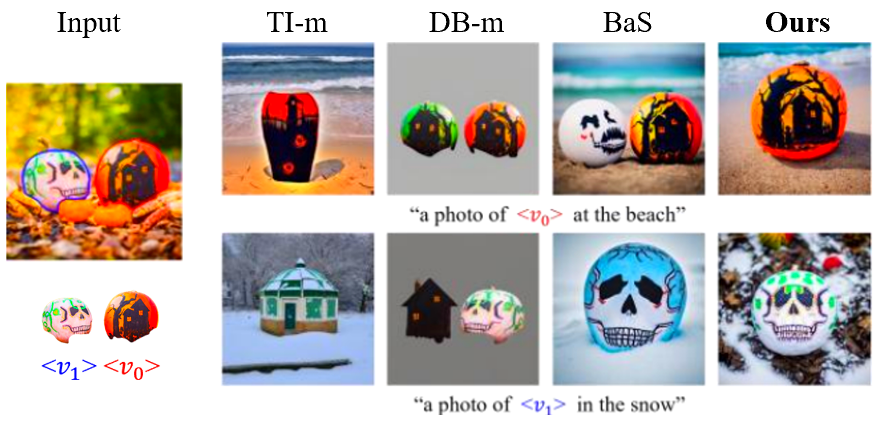}
    \includegraphics[width=1\linewidth]{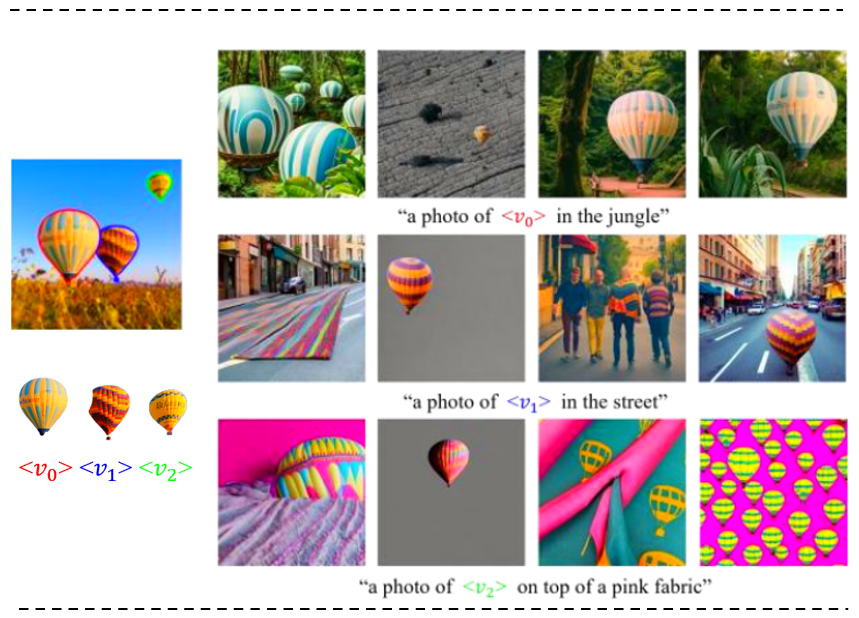}
    \includegraphics[width=1\linewidth]{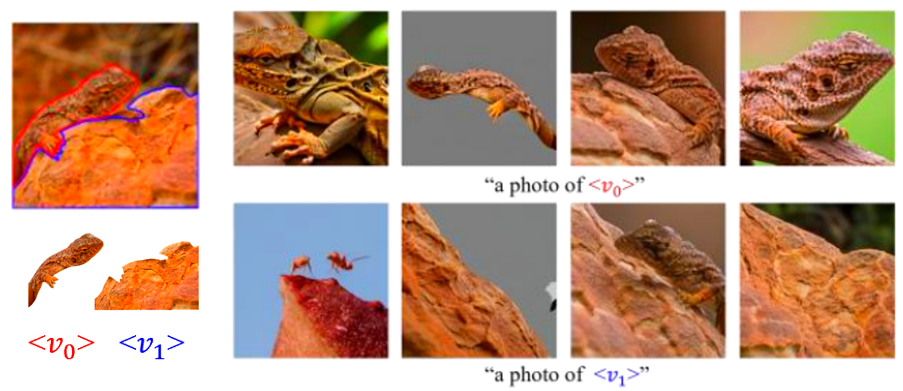}
    \caption{
        \textbf{More qualitative comparison of semantic learning. }
    }
    \label{fig-more_qualitative_comparison_semantic_learning}
\end{figure}

Fig.~\ref{fig-more_qualitative_comparison_semantic_learning} presents more qualitative comparison against baseline methods. 

\begin{figure}[htbp]
    \centering
    \includegraphics[width=1\linewidth]{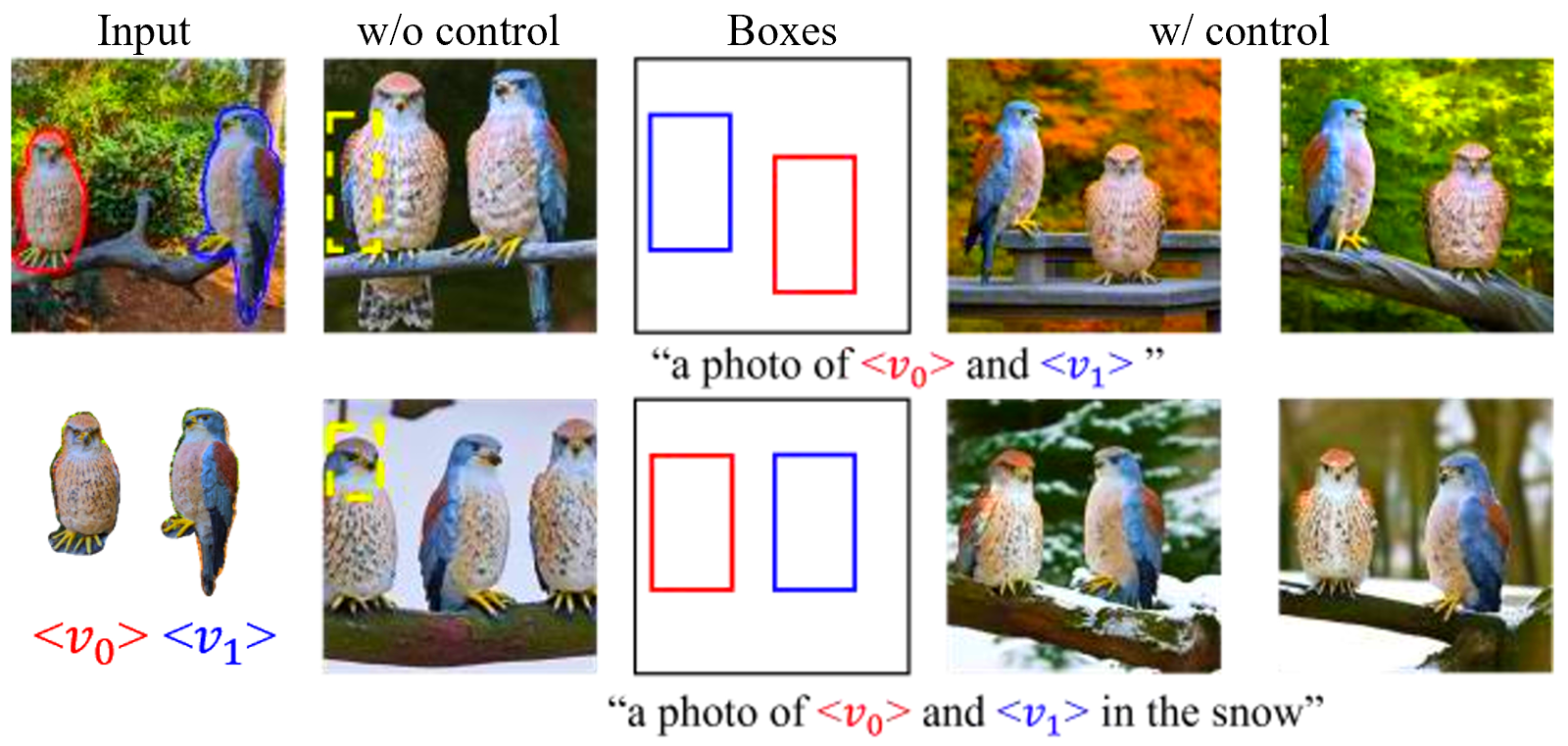}
    \caption{
        \textbf{Qualitative results of precise synthesis with semantically similar objects.}
        Semantic leakage are marked with yellow dashed boxes. 
    }
    \label{fig-mult_obj_synthesis_result_show_semantic}
\end{figure}

\begin{figure}[htbp]
    \centering
    \includegraphics[width=1\linewidth]{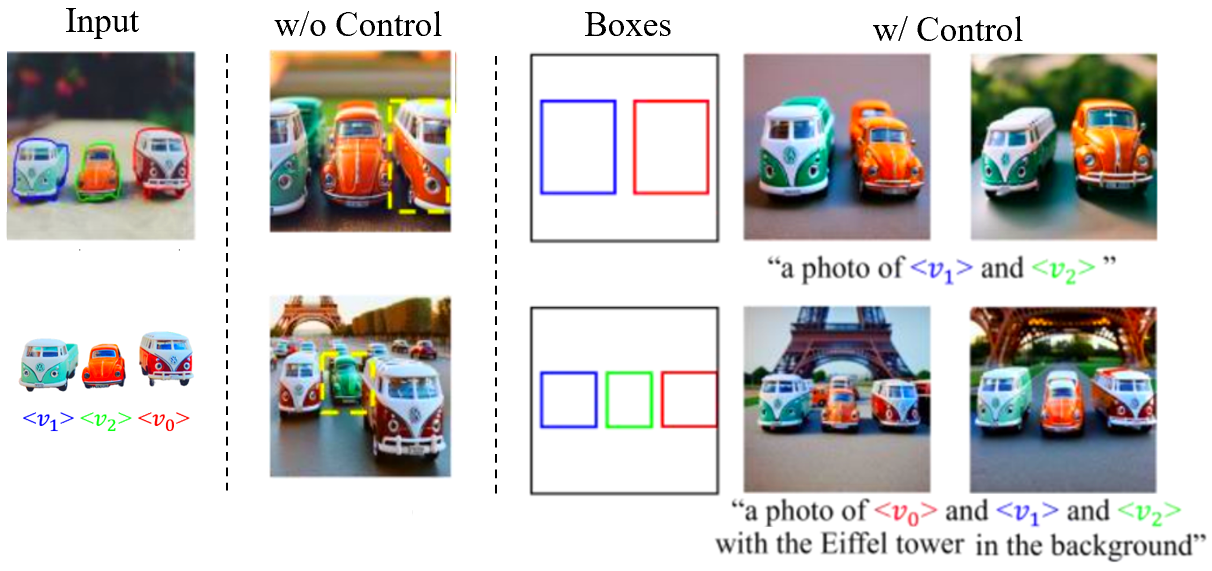}
    \caption{
        \textbf{Qualitative results of precise synthesis with rare-seen objects. }
        Semantic leakage are marked with yellow dashed boxes. 
    }
    \label{fig-mult_obj_synthesis_result_show_rare}
\end{figure}

\paragraph{Precise Synthesis. }
More qualitative results, such as images with semantic similarities or rare-seen objects are presented in Fig.~\ref{fig-mult_obj_synthesis_result_show_semantic} and Fig.~\ref{fig-mult_obj_synthesis_result_show_rare}, respectively. 

\begin{figure}[htbp]
    \centering
    \includegraphics[width=1\linewidth]{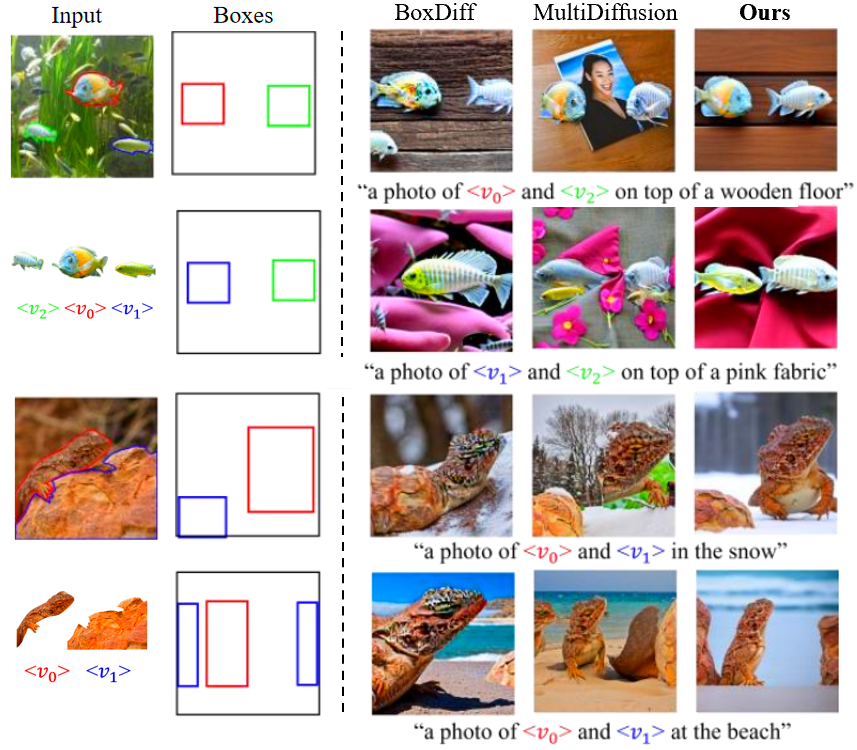}
    \caption{
        \textbf{More qualitative comparison of precise synthesis. }
    }
    \label{fig-more_qualitative_comparison_semantic_synthesis}
\end{figure}

Fig.~\ref{fig-more_qualitative_comparison_semantic_synthesis} presents more qualitative comparison against baseline methods. 


\section{Limitations and Future Work}
\label{Limitations}

We point out the following limitations for future research: 
\begin{enumerate}
    \item Our method requires user-provided masks for semantic segmentation. 
        Future work could explore automatic object grounding and segmentation through the inherent clustering behavior of self-attention mechanisms; 
   \item Our method encodes the semantics of a target instance with a single text embedding, which is constrained by the representation capacity of one token. 
      As a result, the token cannot fully capture the semantics of its corresponding instance, making our approach incompatible with prompt optimization methods such as Promptist that aim to enhance image quality. 
      Future work may explore representing individual instances with multiple tokens.

    \item Our method is based on self-attention control, which may suffer performance degradation under complex prompts due to dispersed attention that weakens focus on background and instance reconstruction. Future work could investigate more robust control mechanisms to improve performance in such scenarios.

\end{enumerate}



\section{Comparison of Reward-Based and Penalty-Based Loss Functions}

We consider an image containing $n$ pixel queries and $K$ learnable instance tokens, along with one background token. Let the attention distribution for pixel $p$ be denoted by
\[
  \mathbf{a}_p = (a_{p0}, a_{p1}, \dots, a_{pK})^\top \in [0,1]^{K+1} 
\]
\[
   \quad\sum_{i=0}^{K} a_{pi} = 1,\quad \forall p \in \{1,\dots,n\},
\]
where $a_{p0}$ corresponds to the background token (aggregating uncontrolled semantics), and $a_{pi}$ for $i \geq 1$ correspond to instance tokens. Each instance $i$ is associated with a binary mask $m_{pi} \in \{0,1\}$, such that $m_{pi}=1$ if and only if pixel $p$ belongs to instance $i$. Each pixel belongs to at most one instance, i.e., $S_p := \sum_{i=1}^{K} m_{pi} \in \{0,1\}$.

\subsection{Reward-Based Loss}

We define the reward-based loss as Eq.~\ref{reward-base_loss_funtion}. Since the layers are independent in calculate attention map, so for each layer the loss function is equivalent to:

\begin{equation}
  \mathcal{L}_{\mathrm{r}} =
  \sum_{p=1}^{n}\sum_{i=1}^{K}
  \left(a_{pi}-\alpha m_{pi}\right)^2, \quad
  \alpha \in (0,1]
  \label{eq:reward-loss}
\end{equation}

\paragraph{Constrained Optimization.}
To ensure attention normalization, we introduce Lagrange multipliers $\lambda_p$:
\[
  \mathcal{J}_{\mathrm{r}} = \mathcal{L}_{\mathrm{r}} +
  \sum_{p=1}^{n}\lambda_p \left(\sum_{i=0}^{K} a_{pi} - 1\right)
\]

\paragraph{First-Order Optimality.}
The partial derivatives of $\mathcal{J}_{\mathrm{r}}$ with respect to $a_{pi}$ yield:
\[
  \frac{\partial\mathcal{J}_{\mathrm{r}}}{\partial a_{pi}} =
  \begin{cases}
    2(a_{pi} - \alpha m_{pi}) + \lambda_p = 0, & i \ge 1,\\
    \lambda_p = 0, & i = 0
  \end{cases}
\]
Solving gives
\begin{equation}
  a_{pi} =
  \begin{cases}
    \alpha m_{pi} - \dfrac{\lambda_p}{2}, & i \ge 1,\\[6pt]
    1 -\sum_{i=1}^Ka_{pi}, & i = 0
  \end{cases}
  \label{eq:reward-solution}
\end{equation}

Since our analysis focuses exclusively on the case where $i \neq 0$we disregard the case $i = 0$the following discussion.

\paragraph{Solving \texorpdfstring{$\lambda_p$}{lambda}.}
Define $S_{p}=\sum_{i=1}^K m_{pi}$ and substituting into the normalization constraint:
\[
  \alpha S_p - \frac{(K+1)\lambda_p}{2} = 1
  \quad\Rightarrow\quad
  \lambda_p = \frac{2\alpha S_p - 2}{K+1}
\]

\paragraph{Interpretation.} For Instance pixels ($S_p = 1$): $\lambda_p = \frac{2\alpha - 2}{K+1} < 0$ (for $\alpha < 1$).\\
  Suppose pixel $p$ belongs to instance $j$, then:
  \[
    a_{pj} = \alpha - \frac{\lambda_p}{2}, \quad
    a_{pi \ne j} = -\frac{\lambda_p}{2} > 0
  \]
  Due to $\alpha$ be set as $\frac{1}{2}$ so that:
  \[
    \lambda_p = \frac{-1}{K+1} \quad\Rightarrow\quad
    a_{pi \ne j} = \frac{1}{2(K+1)} > 0
  \]
This implies that non-target tokens receive non-zero attention, resulting in semantic entanglement and non-unique solutions.

\paragraph{Conclusion.}
Reward-based loss aligns attention with relevant semantics but fails to penalize irrelevant activations, leading to flat solution landscapes and entangled attention distribution.

\subsection{Penalty-Based Loss}

As same as reward-based loss function, the penalty-base loss function(Eq.~\ref{penalty-base_loss_function}) is equivalent to:
\begin{equation}
  \mathcal{L}_{\mathrm{p}} =
  \sum_{p=1}^{n}\sum_{i=1}^{K}
  (1 - m_{pi})\,a_{pi}^2
  \label{eq:penalty-loss}
\end{equation}

\paragraph{Constrained Optimization.}
\[
  \mathcal{J}_{\mathrm{p}} = \mathcal{L}_{\mathrm{p}} +
  \sum_{p=1}^{n} \mu_p \left(\sum_{i=0}^{K} a_{pi} - 1\right)
\]

\paragraph{First-Order Optimality.}
\[
  \frac{\partial\mathcal{J}_{\mathrm{p}}}{\partial a_{pi}} =
  \begin{cases}
    2(1 - m_{pi})\,a_{pi} + \mu_p = 0, & i \ge 1,\\
    \mu_p = 0, & i = 0
  \end{cases}
\]

\paragraph{Case Analysis.}

As the problem fulfills the Karush–Kuhn–Tucker (KKT) conditions, we can proceed with the following analysis.

\begin{itemize}
  \item \textbf{If pixel $p$ belongs to instance $j$}: $m_{pj}=1$, $m_{pi}=0$ for $i \ne j$ $\Rightarrow \mu_p = 0$, so exists a theoretical optimal solution:
  \[
    a_{pj} = 1,\quad a_{pi \ne j} = 0,\quad a_{p0} = 0
  \]
\end{itemize}

\paragraph{Convexity and Uniqueness.}
The Hessian is diagonal with
\[
  \frac{\partial^2\mathcal{L}_{\mathrm{p}}}{\partial a_{pi}^2} = 2(1 - m_{pi}) \ge 0,
\]
and strictly positive for $i$ with $m_{pi} = 0$. Thus, the optimization is convex and admits a unique global solution.

\paragraph{Conclusion.}
Penalty-based loss suppresses attention to irrelevant semantics, ensuring uniqueness and disentanglement of attention. This yields semantically clean, interpretable, and spatially localized token distributions.

\end{document}